\documentclass{article}

\usepackage{PRIMEarxiv}

\usepackage[utf8]{inputenc} 
\usepackage[T1]{fontenc}    
\usepackage{hyperref}       
\usepackage{url}            
\usepackage{booktabs}       
\usepackage{amsfonts}       
\usepackage{nicefrac}       
\usepackage{microtype}      
\usepackage{lipsum}
\usepackage{fancyhdr}       
\usepackage{graphicx}       
\usepackage[linesnumbered,ruled,vlined]{algorithm2e}
\graphicspath{{media/}}     
\usepackage{subcaption}
\usepackage{amsmath}
\usepackage{epstopdf}
\usepackage{amsthm}

\newtheorem{definition}{Definition}
\newtheorem{assumption}{Assumption}

\title{NeuroSep-CP-LCB: A Deep Learning-based Contextual Multi-armed Bandit Algorithm with Uncertainty Quantification for Early Sepsis Prediction
}

\author{
  Anni Zhou, Raheem Beyah \\
  School of Electrical and Computer Engineering\\
  Georgia Institute of Technology \\
  \texttt{\{azhou60, ab207\}@gatech.edu} \\
   \And
  Rishikesan Kamaleswaran \\
  School of Medicine, Department of Surgery\\
  Duke University \\
  \texttt{r.kamaleswaran@duke.edu} 
}

\begin{document}
\maketitle

\begin{abstract}
In critical care settings, timely and accurate predictions can significantly impact patient outcomes, especially for conditions like sepsis, where early intervention is crucial. We aim to model patient-specific reward functions in a contextual multi-armed bandit setting. The goal is to leverage patient-specific clinical features to optimize decision-making under uncertainty.
In this paper, we propose NeuroSep-CP-LCB, a novel integration of neural networks with contextual bandits and conformal prediction tailored for early sepsis detection. Unlike the algorithm pool selection problem in the previous paper, where the primary focus was identifying the most suitable pre-trained model for prediction tasks, this work directly models the reward function using a neural network, allowing for personalized and adaptive decision-making. Combining the representational power of neural networks with the robustness of conformal prediction intervals, this framework explicitly accounts for uncertainty in offline data distributions and provides actionable confidence bounds on predictions.
\end{abstract}


 \section{Introduction}

The study of offline policy learning (OPL) has seen significant advancements in both bandit and reinforcement learning settings. Traditional approaches often rely on tabular representations \cite{yin2020asymptotically, rashidinejad2021bridging} or linear function approximations \cite{duan2020minimax,jin2021pessimism}. These works leverage analytical optimization or oracle-based methods to derive sub-optimality guarantees, albeit under stringent data coverage assumptions and limited scalability. While these approaches provide theoretical insights, their practical application is hindered by challenges in handling high-dimensional or complex data distributions.

Recent works have explored general function approximators, such as neural networks, for OPL \cite{nguyen2021sample,david2021offline}. Neural-based approaches show promise due to their representational power, particularly when paired with techniques such as stochastic gradient descent for optimization. Notably, Nguyen-Tang et al. \cite{nguyen2021sample} utilize narrow neural networks with strong uniform data coverage assumptions, achieving sub-optimality bounds under specific oracle conditions. However, these methods often impose constraints that may limit their generalization to real-world scenarios with non-uniform or incomplete data coverage.


The adoption of neural networks in policy learning has been inspired by their empirical success and theoretical properties, including their ability to approximate overparameterized models while generalizing effectively \cite{allen2019convergence,cao2019generalization}. Within this domain, \textbf{NeuraLCB} \cite{nguyen-tang2022offline} stands out as a key development. It is tailored for online contextual bandits and employs neural networks to construct confidence bounds for decision-making. By iteratively updating the neural model in response to streaming data, NeuraLCB balances computational efficiency with theoretical guarantees. Despite these advances, NeuraLCB's reliance on streaming updates and its application to online settings leave gaps in its applicability to the offline policy learning context, where data is static, and robustness to distributional shifts is critical.

Conformal prediction has emerged as a powerful framework for quantifying uncertainty in machine learning models \cite{conformaltutorial,angelopoulos2021gentle}. Its ability to provide distribution-free, finite-sample guarantees makes it particularly appealing for applications requiring reliable uncertainty estimation. While conformal methods have been widely applied in supervised learning contexts, their integration into policy learning remains under-explored. Recent studies have begun to investigate the potential of conformal approaches to address issues such as covariate shift \cite{tibshirani2019conformal} and sparse coverage in offline settings \cite{lei2021conformal}, suggesting that combining conformal prediction with advanced function approximators may offer robust guarantees for OPL.

 
 
In this paper, building on the strengths of neural networks, contextual bandit and conformal prediction, we proposed a framework that integrates these methodologies to address key challenges in offline policy learning. By leveraging NeuraLCB's over-parameterized neural architecture for expressive policy representation and conformal prediction's robust uncertainty quantification, our approach improves generalization under weak or non-uniform data coverage conditions. Unlike traditional neural-based OPL methods, our framework explicitly accounts for uncertainty in offline data distributions, providing calibrated confidence intervals and reducing sub-optimality. This novel synthesis of tools establishes a new OPL state-of-the-art, offering theoretical rigor and practical scalability. 
The key contributions of this paper can be summarized as follows:
\begin{itemize}
    \item \textbf{Reward Function Approximation with Neural Networks:} NeuroSep-CP-LCB directly approximates the reward function from raw patient data, enabling granular and adaptive decision-making.
 \item \textbf{Integrated Conformal Prediction for Uncertainty Quantification:} The conformal prediction intervals provide calibrated uncertainty bounds for clinical decision-making and guide the exploration-exploitation trade-off in action selection policies.
\item \textbf{Patient-specific and Action-aware Decisions:} By utilizing the prediction intervals in the policy update step, the algorithm ensures decisions are both context-aware and robust to data variability.
\end{itemize}
This paper addresses the pressing challenge of optimizing decisions in high-stakes medical scenarios where uncertainty is intrinsic, data coverage is incomplete, and errors can have severe consequences. Through a principled integration of neural networks, contextual bandits, and conformal prediction, NeuroSep-CP-LCB offers a state-of-the-art solution that extends beyond mere model selection to improve patient outcomes directly.

The remainder of the paper is organized as follows: Section \ref{sec:mat} demonstrates the details of the problem formulation and mathematical foundation of the proposed framework. Section \ref{sec:res} provides the details of the dataset description, data preprocessing methods, experimental setup, and experimental results, and Section \ref{sec:disc} discusses and analyzes the results. Finally, we conclude the study in Section \ref{sec:conc}.

\section{Materials and methods}
\label{sec:mat}
\subsection{Problem Formulation}
Following similar steps in \cite{nguyen-tang2022offline}, we model the early sepsis prediction problem as a stochastic contextual bandits problem and utilize the over-parameterized neural network to approximate the unknown reward function. Since our neural network is built on \cite{nguyen-tang2022offline}, we use the same notations as that in \cite{nguyen-tang2022offline} for clarity and simplicity.


Considering a stochastic \(K\)-armed contextual bandit, at each round \(t\), we have an online learner who observes a full context \(x_t:= \{x_{t, a} \in \mathbb{R}^d: a \in [K]\}\) sampled from a distribution \(\rho\), where \([K]\) denotes the finite action space. At each round \(t\), the online learner takes an action \(a_t\in[K]\), and receives a reward \(r_t \sim P(\cdot | x_{t,a_t})\). We denote \(\pi\) as a policy that maps a context to a distribution over the action space \([K]\). For each context \(x:= \{x_a \in \mathbb{R}^d: a \in [K]\}\), we define \(v^{\pi}(x) = \mathbb{E}_{a \sim \pi(\cdot | x), r \sim P(\cdot | x_a)}[r]\) and \(v^*(x) = \max_{\pi} v^{\pi}(x)\). Since we have a finite action space, \(v^*(x)\) is attainable. 
The performance of a learned policy \(\hat{\pi}\) is measured by the expected sub-optimality achieved across the unknown context distribution \(\rho\):
\begin{equation}\label{eq: subopt_pi}
    SubOpt(\hat{\pi}) := \mathbb{E}_{x \sim \rho} [ SubOpt (\hat{\pi}; x)], \quad \text{where} \quad SubOpt (\hat{\pi}; x) := v^*(x) - v^{\hat{\pi}}(x).
\end{equation}

As in a standard offline contextual bandit problem \cite{nguyen-tang2022offline}, our goal is to learn an optimal policy only from the offline dataset \(\mathcal{D}_n = \{(x_t, a_t, r_t)\}_{t=1}^n\), which is collected prior by a behavior policy \(\mu\).

\begin{assumption}
(\(R\)-subgaussian noise assumption) Following similar steps in previous work \cite{abbasi2011improved,zhou2020neural,xiao2021optimality}, we denote \(h : \mathbb{R}^d \to [0, 1]\) as an unknown reward function and \(\xi_t\) as a \(R\) sub-Gaussian noise conditioned on \((\mathcal{D}_{t-1}, x_t, a_t)\) where \(\mathcal{D}_t = \{(x_{\tau}, a_{\tau}, r_{\tau})\}_{1 \leq \tau \leq t}\), \(\forall t\). Then, we have:
\begin{equation}\label{eq: assump1}
    r_t = h(x_{t,a_t}) + \xi_t, \forall t. 
\end{equation}
\end{assumption}

The \(R\)-subgaussian noise assumption is standard in stochastic bandit literature and is satisfied for any bounded noise \cite{nguyen-tang2022offline}.


Then, we approximate the unknown reward function via a neural network, and the reward function is without prior knowledge of its parametric form.
We use an over-parameterized neural network to learn the unknown reward function, i.e., the width $m$ of the neural network is sufficiently larger than the number of samples $n$.
The fully connected neural network is considered with depth \(L \geq 2\) defined in \(\mathbb{R}^d\) as
\begin{equation}\label{eq:f_W(u)}
f_{\boldsymbol{W}}(\boldsymbol{u}) = \sqrt{m} \boldsymbol{W}_L \sigma (\boldsymbol{W}_{L-1} \sigma (\dots \sigma (\boldsymbol{W}_1 \boldsymbol{u}) \dots)), \forall \boldsymbol{u} \in \mathbb{R}^d,
\end{equation}
in which $\sigma(\cdot) = \max \{ \cdot, 0 \}$ is the rectified linear unit (ReLU) activation function.
We have 
\begin{equation}\label{eq:WL}
    \boldsymbol{W}_1 \in \mathbb{R}^{m \times d};
    \boldsymbol{W}_i \in \mathbb{R}^{m \times m}, \forall i \in [2, L-1], \boldsymbol{W}_L \in \mathbb{R}^{m \times 1};
    \boldsymbol{W} := (\boldsymbol{W}_1, \dots, \boldsymbol{W}_L), \text{vec}(\boldsymbol{W}) \in \mathbb{R}^p,
\end{equation}
where $p = md + m + m^2(L-2)$. 
Under such a regime, the training dynamics of the neural network can be captured in the framework of the so-called neural tangent kernel (NTK) \cite{jacot2018neuralNTK}. 
Previous works have shown that overparameterization is effective in studying neural training and the interpolation phenomenon for deep neural networks \cite{nguyen-tang2022offline, belkin2021fit, allen2019convergence, hanin2019finite, arora2019exact, cao2019generalization}.

Our study relies on the neural tangent kernel (NTK) introduced by \cite{jacot2018neural}. As in \cite{nguyen-tang2022offline}, we define the NTK matrix for the neural network function in \ref{eq:f_W(u)} as follows:
\begin{definition}
[Neural Tangent Kernel Matrix \cite{jacot2018neural,cao2019generalization,zhou2020neural,nguyen-tang2022offline}] 
Let 
\[
\{\boldsymbol{x}^{(i)}\}_{i=1}^{nK} = \{x_{t,a} \in \mathbb{R}^d: t \in [n], a \in [K]\}, \widehat{\boldsymbol{H}}_{i,j}^{(l)} = \Sigma_{i,j}^{l+1}=\langle\boldsymbol{x}^{(i)},\boldsymbol{x}^{(j)}\rangle, \] 
\[
\boldsymbol{A}_{i,j}^{(l)} = 
\begin{bmatrix}
\boldsymbol{\Sigma}_{i,i}^{(l)} & \boldsymbol{\Sigma}_{i,j}^{(l)} \\
\boldsymbol{\Sigma}_{i,j}^{(l)} & \boldsymbol{\Sigma}_{j,j}^{(l)}
\end{bmatrix},
\]
\[
\boldsymbol{\Sigma}_{i,j}^{(l+1)} = 2 \mathbb{E}_{(u,v) \sim \mathcal{N}(0, \boldsymbol{A}_{i,j}^{(l)})} \left[ \sigma(u) \sigma(v) \right],
\]
\[ 
\widehat{\boldsymbol{H}}_{i,j}^{(l+1)}=2\widehat{\boldsymbol{H}}_{i,j}^{(l)}\mathbb{E}_{(u,v) \sim \mathcal{N}(0, \boldsymbol{A}_{i,j}^{(l)})} \left[ \sigma'(u) \sigma'(v) \right] 
+  \boldsymbol{\Sigma}{i,j}^{(l+1)}.
\]
\text{Then, the NTK matrix is defined as}
\[\boldsymbol{H}= (\widehat{\boldsymbol{H}}^{(L)} + \boldsymbol{\Sigma}^{(L)}) / 2.\]
\end{definition}
The Gram matrix $\boldsymbol{H}$ is derived recursively, starting from the first layer and proceeding to the final layer of the neural network, by employing Gaussian distributions over the observed contexts $\{\boldsymbol{x}^{(i)}\}_{i=1}^{nK}$. We then outline the assumptions commonly made in overparameterized neural network literature. 

\begin{assumption}
[Non-singularity \cite{arora2019exact, du2018gradient, du2019gradient, cao2019generalization}]
$\exists \lambda_0 > 0, \boldsymbol{H} \succeq \lambda_0 \boldsymbol{I}, \text{ and } \forall i \in [nK], \|x^{(i)}\|_2 \leq 1$ and $[x^{(i)}]_j = [x^{(i)}]_{j+\frac{d}{2}} \, \forall i \in [nK], j  \in [d/2].$
\label{ass:n1}
\end{assumption}
The Non-singularity assumption states that $\boldsymbol{H}$ is non-singular and that the input data lies within the unit sphere $\mathbb{S}^{d-1}$. It is frequently used in the literature on overparameterized neural networks \cite{arora2019exact, du2018gradient, du2019gradient, cao2019generalization}. Non-singularity is satisfied when any two contexts $\{x^{(i)}\}$ are not parallel \cite{zhou2020neural}. The unit sphere condition simplifies the scope of the analysis and can be extended to cases where the input data is bounded in 2-norm. Additionally, for any input data point $\boldsymbol{x}$ such that $\|\boldsymbol{x}\|_2 = 1$, we can always construct a new input $\boldsymbol{x}' = \frac{1}{2}[\boldsymbol{x}, \boldsymbol{x}^{\top}]$.  
Specifically, under the Non-singularity assumption and the initialization scheme in \ref{alg:neuralcb_bmode}, we have $f_{\boldsymbol{W}^0}(\boldsymbol{x}^{(i)}) = 0, \forall i \in [nK]$.

\subsection{Neural Sepsis Prediction with Conformal Prediction and LCB (NeuroSep-CP-LCB)}

\begin{figure}[!ht]
    \centering
    \includegraphics[width=\linewidth]{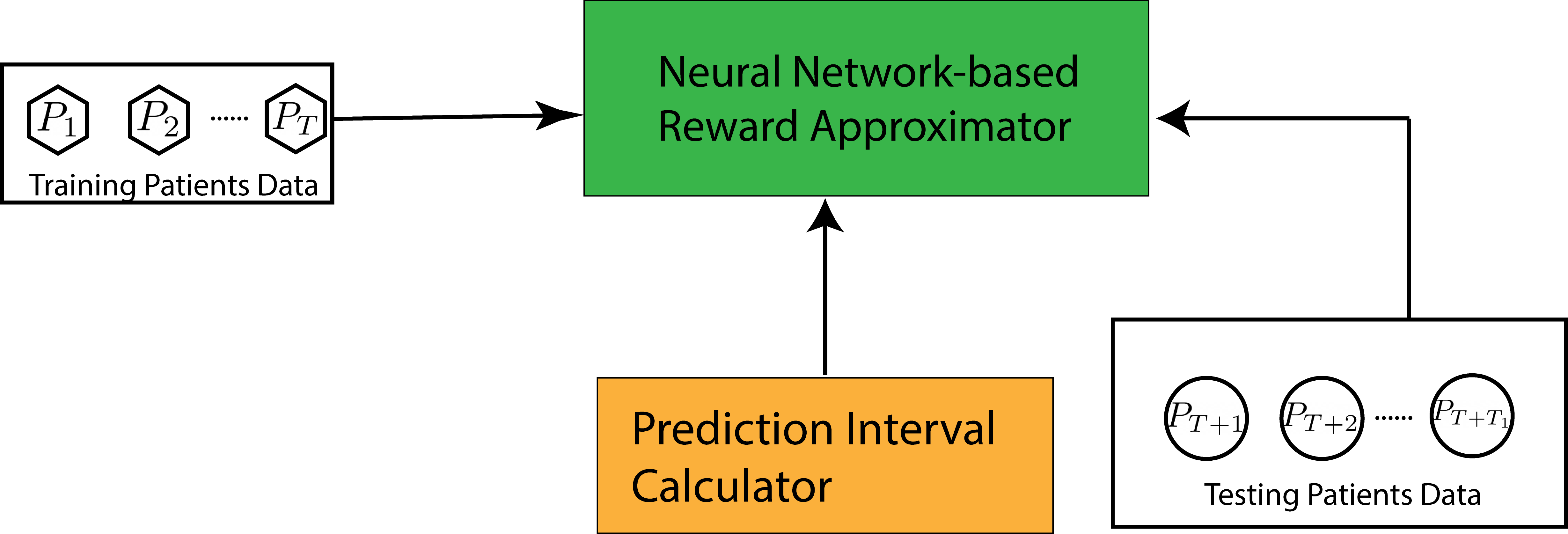}
    \caption{System Overview.}
    \label{fig:neuralcbcp_highlevel}
\end{figure}

NeuroSep-CP-LCB is particularly designed for early sepsis detection of ICU patients, leveraging a contextual bandit framework integrated with conformal prediction to generate reliable prediction intervals of each prediction hourly. The proposed algorithm enables the careful selection of actions (i.e., predicting a non-septic (class 0) or septic (class 1) state) based on a patient’s clinical context. The algorithm operates on an offline dataset $\mathcal{D}_n = \{(\boldsymbol{x}_t, a_t, r_t)\}_{t=1}^n$. Given the round $t$, $\boldsymbol{x}_t$ represents the patient's clinical features (i.e., contexts), $a_t$ is the selected action, and $r_t$ is the observed reward. A high-level system overview is shown in \ref{fig:neuralcbcp_highlevel}.

\begin{algorithm}[!ht]
\caption{NeuroSep-CP-LCB.}
\label{alg:neuralcb_bmode}
\KwIn{Offline dataset $\mathcal{D}_n = \{(\boldsymbol{x}_t, a_t, r_t)\}_{t=1}^n$; $\{\mathcal{D}_n' =(x_j, a_j, y_j)\}_{j=n+1}^{n+n_1}$; step sizes $\{\eta_t\}_{t=1}^n$; regularization parameter $\lambda > 0$; confidence parameters $\{\beta_t\}_{t=1}^n$; $B > 0$; epoch number $J > 0$.}
\SetKwInOut{Require}{Require}
\SetKwInOut{Output}{Output}
\Require{
$\boldsymbol{W}_l^{(0)} = [\boldsymbol{\widetilde{W}_l}; 0; 0; \boldsymbol{\widetilde{W}_l}], \forall l \in [L-1]$ in which each entry of $\boldsymbol{\widetilde{W}_l}$ is independently sampled from $\mathcal{N}(0, 4/m)$, and set $\boldsymbol{W}_L^{(0)} = [\boldsymbol{w}^T, -\boldsymbol{w}^T]$ where each entry of $\boldsymbol{w}$ is sampled from $\mathcal{N}(0, 2/m)$.}
$\boldsymbol{\Lambda}_0 \gets \lambda \boldsymbol{I}$\;
\For{$t = 1, \dots, n$}{
    Retrieve $(\boldsymbol{x}_t, a_t, r_t)$ from $\mathcal{D}_n$\;
    Get predict intervals and conformal prediction estimators $\{ C_j^{\phi, \alpha}(\boldsymbol{u})|_{j = n+1}^{n+n_1}, f^B_{\boldsymbol{W}^{(t-1)}}(\boldsymbol{u}) \} \gets EnsembleCP1 (\mathcal{D}_n,\mathcal{D}_n', \mathcal{F}, B) $\;
    $L_t(\boldsymbol{u}) \gets f^B_{\boldsymbol{W}^{(t-1)}}(\boldsymbol{u}) - \beta_t \| \nabla f_{\boldsymbol{W}^{(t-1)}}(\boldsymbol{u}) \|_{\boldsymbol{\Lambda}_{t-1}^{-1}}, \forall \boldsymbol{u} \in \mathbb{R}^d$\;
    $\hat{\pi}_t(\boldsymbol{x}) \gets \arg\max_{a \in [K]} L_t(\boldsymbol{x}_t, a_t), \forall  \boldsymbol{x}_t \in \mathbb{R}^d , a_t \in [K]$\;
    $\boldsymbol{\Lambda}_t \gets \boldsymbol{\Lambda}_{t-1} + \mathrm{vec}(\nabla f_{\boldsymbol{W}^{(t-1)}}(\boldsymbol{x}_t, a_t)) \cdot \mathrm{vec}(\nabla f_{\boldsymbol{W}^{(t-1)}}(\boldsymbol{x}_t, a_t))^T / m$\;
    $\boldsymbol{\widetilde{W}}^{(0)} \gets \boldsymbol{W}^{(t-1)}$\;
    
    \For{$j = 1, \dots, J$}{
        Sample a batch of data $B_t = \{(\boldsymbol{x}_{t_q}, a_{t_q}, r_{t_q})\}_{q=1}^B$ from $\mathcal{D}_t$\;
        $\mathcal{L}^{(j)}_t(\boldsymbol{W}) \gets \frac{1}{2B} \sum_{q=1}^B (f_{\boldsymbol{W}}(\boldsymbol{x}_{t_q}, a_{t_q}) - r_{t_q})^2 + \frac{m \lambda}{2} \|\boldsymbol{W} - \boldsymbol{W}^{(0)}\|_F^2$\;
        $\boldsymbol{\widetilde{W}}^{(j)} \gets \boldsymbol{\widetilde{W}}^{(j-1)} - \eta_t \nabla \mathcal{L}_t^{(j)}(\boldsymbol{\widetilde{W}}^{(j-1)})$\;
    }  
    $\boldsymbol{W}^{(t)} \gets \boldsymbol{\widetilde{W}}^{(J)}$\;
}
\Output{Randomly sample $\hat{\pi}$ uniformly from $\{\hat{\pi}_1, \dots, \hat{\pi}_n\}, C_j^{\phi, \alpha}(\boldsymbol{u}).$}
\end{algorithm}

Following the same steps as that in \cite{nguyen-tang2022offline}, the proposed algorithm initializes the neural network parameters $\boldsymbol{W}_l^{(0)}$ for each layer $l$ using Gaussian distributions to ensure appropriate variability. Specifically, each entry of $\boldsymbol{\widetilde{W}_l}$ is sampled from $\mathcal{N}(0, 4/m)$, while the final layer parameters $\boldsymbol{W}_L^{(0)}$ are initialized using a scaled Gaussian distribution $\mathcal{N}(0, 2/m)$. The confidence matrix $\boldsymbol{\Lambda}_0$ is then initialized as a regularized identity matrix, $\boldsymbol{\Lambda}_0 \gets \lambda \boldsymbol{I}$, where $\lambda > 0$ is a regularization parameter.

At each time step $t = 1, \dots, n$, the algorithm retrieves the context $\boldsymbol{x}_t$, action $a_t$, and reward $r_t$ from the clinical dataset $\mathcal{D}_n$. The algorithm invokes the \texttt{EnsembleCP1} subroutine, which constructs ensemble-based conformal prediction intervals to quantify uncertainty. The LCB for each context-action pair $\boldsymbol{u} \in \mathbb{R}^d$ is then computed as
\begin{equation}
L_t(\boldsymbol{u}) \gets f^B_{\boldsymbol{W}^{(t-1)}}(\boldsymbol{u}) - \beta_t \| \nabla f_{\boldsymbol{W}^{(t-1)}}(\boldsymbol{u}) \|_{\boldsymbol{\Lambda}_{t-1}^{-1}},
\end{equation}
where $f^B_{\boldsymbol{W}^{(t-1)}}$ denotes the neural network output, $\nabla f_{\boldsymbol{W}^{(t-1)}}$ is the gradient, and $\beta_t$ is a confidence parameter. The action $\hat{\pi}_t(\boldsymbol{x})$ that maximizes the LCB is selected using
\begin{equation}
\hat{\pi}_t(\boldsymbol{x}) \gets \arg\max_{a \in [K]} L_t(\boldsymbol{x}_t, a_t), \forall  \boldsymbol{x}_t \in \mathbb{R}^d , a_t \in [K],
\end{equation}
where $K = 2$, corresponding to the non-septic and septic classes.

The confidence matrix $\boldsymbol{\Lambda}_t$ is updated to reflect new data using
\begin{equation}
\boldsymbol{\Lambda}_t \gets \boldsymbol{\Lambda}_{t-1} + \frac{\mathrm{vec}(\nabla f_{\boldsymbol{W}^{(t-1)}}(\boldsymbol{x}_t, a_t)) \cdot \mathrm{vec}(\nabla f_{\boldsymbol{W}^{(t-1)}}(\boldsymbol{x}_t, a_t))^T}{m},
\end{equation}
where $m$ is a normalization factor. The neural network parameters are optimized via stochastic gradient descent (SGD) over multiple epochs. The loss function for each epoch is given by
\begin{equation}
\mathcal{L}^{(j)}_t(\boldsymbol{W}) = \frac{1}{2B} \sum_{q=1}^B \left(f_{\boldsymbol{W}}(\boldsymbol{x}_{t_q}, a_{t_q}) - r_{t_q}\right)^2 + \frac{m \lambda}{2} \|\boldsymbol{W} - \boldsymbol{W}^{(0)}\|_F^2,
\end{equation}
where $B$ is the number of bootstrap models in conformal prediction, $\lambda$ is the regularization parameter, and $\boldsymbol{W}^{(0)}$ are the initial weights. Finally, a policy $\hat{\pi}$ is randomly sampled from the set $\{\hat{\pi}_1, \dots, \hat{\pi}_n\}$, and the calibrated prediction intervals $C_j^{\phi, \alpha}(x_j)$ are outputted.

\begin{algorithm}[!ht]
\caption{EnsembleCP1 ($\mathcal{D}_n,\mathcal{D}_n', \mathcal{F}, B$).}
\label{algo:EnsembleCP1}
\SetKwInOut{Require}{Require}
\SetKwInOut{Output}{Output}
\SetKwInOut{Initialize}{Initialize}
\Require{Offline training dataset $\mathcal{D}_n = \{(x_i, a_i, r_i)\}_{i=1}^n$; neural sepsis prediction algorithm $\mathcal{F}$; significance level $\alpha$; aggregation function $\phi$; the number of bootstrap models $B$; 
batch size $bs$; 
testing dataset $\{\mathcal{D}_n' =(x_j, a_j, y_j)\}_{j=n+1}^{n+n_1}$; $y_j$ and $a_j$ are revealed at the end of round $j$ after prediction is finished.}
\Initialize{$\boldsymbol{\hat{\epsilon}}  = \{\}$ as an ordered set\;}
\For{$b = 1, \dots, B$}{
    Sample with replacement an index set $S_b = (i_1, \dots, i_T)$ from indices $(1, \dots, T)$\;
    Compute $\hat{f}_b = \mathcal{F} ((x_i, y_i), i \in S_b)$\;
}

\For{$i = 1, \dots, n$}{
    $\hat{f}_{-i}^{\phi}(x_i) = \phi(\hat{f}_b(x_i), i \notin S_b)$\;
    Compute $\boldsymbol{\hat{\epsilon}_i^{\phi}} = y_i -\hat{f}_{-i}^{\phi}(x_i)$\;
    $\boldsymbol{\hat{\epsilon}} = \boldsymbol{\hat{\epsilon}}  \cup \{\boldsymbol{\hat{\epsilon}_i^{\phi}}\}$\;
}
\For{$j = n+1, \dots, n+n_1$}{
    $\hat{q}_{-j}^{\phi }(x_j) = \phi(\hat{f}_{-i}^{\phi}(x_j), i = 1, \dots, n)$\;
     Compute $\boldsymbol{\hat{\epsilon}_j^{\phi}} = y_j -\hat{q}_{-j}^{\phi }(x_j)$\;
     $\boldsymbol{\hat{\epsilon}} = \boldsymbol{\hat{\epsilon}}  \cup \{\boldsymbol{\hat{\epsilon}_j^{\phi}}\}$\;
    $\hat{\beta} = \arg\min_{\beta \in [0, \alpha]}(1 - \alpha + \beta) \text{ quantile of } \boldsymbol{\hat{\epsilon}}  - \beta \text{ quantile of } \boldsymbol{\hat{\epsilon}} $\;
     
    $width\_lower_j^{\phi, \alpha } = \hat{\beta}$ quantile of $\boldsymbol{\hat{\epsilon}}$\;
    $width\_upper_j^{\phi, \alpha } = (1 - \alpha + \hat{\beta})$ quantile of $\boldsymbol{\hat{\epsilon}}$\;
    $C_j^{\phi, \alpha}(x_j) = [\hat{q}_{-j}^{\phi}(x_j) + width\_lower_j^{\phi, \alpha }, \hat{q}_j^{\phi}(x_j) + width\_upper_j^{\phi, \alpha}]$\;
    
    \If{$j - n \mod bs = 0$}{
        \For{$q = j - bs, \dots, j - 1$}{
            Compute $\boldsymbol{\hat{\epsilon}_q^{\phi}} = y_q - \hat{f}_{-q}^{\phi}(x_j)$\;
            $\boldsymbol{\hat{\epsilon}}  = \boldsymbol{\hat{\epsilon}} \cup \{ \boldsymbol{\hat{\epsilon}_q^{\phi}}  \}$ and reset index of $\boldsymbol{\hat{\epsilon}}$\;
        }
    }
}
\Output{ $\{C_j^{\phi, \alpha}(x_j), \hat{q}_{-j}^{\phi }(x_j)\}|_{j = n+1}^{n+n_1}$}
\end{algorithm}
\paragraph{Subroutine EnsembleCP1}
The \texttt{EnsembleCP1} subroutine constructs ensemble-based conformal prediction intervals to assess the reliability of predictions. Given an offline training dataset $\mathcal{D}_n$, a neural sepsis prediction model $\mathcal{F}$, a significance level $\alpha$, and an aggregation function $\phi$, the subroutine first trains $B$ bootstrap models. For each bootstrap model $b = 1, \dots, B$, an index set $S_b$ is sampled with replacement from $(1, \dots, T)$, and the model $\hat{f}_b$ is trained on this sampled data.

The subroutine computes residuals for each training example $i = 1, \dots, n$ using leave-one-out (LOO) estimators, $\hat{f}_{-i}^{\phi}(x_i)$, and calculates $\boldsymbol{\hat{\epsilon}_i^{\phi}} = y_i - \hat{f}_{-i}^{\phi}(x_i)$. These residuals are aggregated into an ordered set $\boldsymbol{\hat{\epsilon}}$. For test points $j = n+1, \dots, n+n_1$, conformal prediction intervals $C_j^{\phi, \alpha}(x_j)$ are constructed using quantiles of $\boldsymbol{\hat{\epsilon}}$, and the intervals are periodically updated using a batch size $bs$.

By providing robust uncertainty estimates through conformal prediction, NeuroSep-CP-LCB supports clinicians in making informed and reliable decisions, enhancing patient safety in critical care settings.

\begin{algorithm}[!ht]
\caption{NeuroSep-CP-LinLCB.}
\KwIn{Offline data $\mathcal{D}_n = \{(\boldsymbol{x}_t, a_t, r_t)\}_{t=1}^n$, $\{\mathcal{D}_n' =(x_j, a_j, y_j)\}_{j=n+1}^{n+n_1}$, regularization parameter $\lambda > 0$, confidence parameters $\beta > 0$, conformal prediction parameter $cpr$, $B$.}
\SetKwInOut{Require}{Require}
 
\Require{
$\boldsymbol{W}_l^{(0)} = [\boldsymbol{\widetilde{W}_l}; 0; 0; \boldsymbol{\widetilde{W}_l}], \forall l \in [L-1]$ in which each entry of $\boldsymbol{\widetilde{W}_l}$ is independently sampled from $\mathcal{N}(0, 4/m)$, and set $\boldsymbol{W}_L^{(0)} = [\boldsymbol{w}^T, -\boldsymbol{w}^T]$ where each entry of $\boldsymbol{w}$ is sampled from $\mathcal{N}(0, 2/m)$.}
$\boldsymbol{\Lambda}_0 \gets \lambda \boldsymbol{I}$\;

$\boldsymbol{\Lambda}_n \leftarrow \lambda \boldsymbol{I} + \sum_{t=1}^{n} \boldsymbol{x}_{t,a_t} \boldsymbol{x}_{t,a_t}^T$\;

$\hat{\boldsymbol{\theta}}_n \leftarrow \boldsymbol{\Lambda}_n^{-1} \sum_{t=1}^{n} \boldsymbol{x}_{t,a_t} r_t$\;

Get the LOO estimator $\{ C_j^{\phi, \alpha}(\boldsymbol{u}), f^B_{\boldsymbol{W}^{(t-1)}}(\boldsymbol{u}) \} \gets EnsembleCP1(\mathcal{D}_n,\mathcal{D}_n', \mathcal{F}, B$)\;

$L(\boldsymbol{u}) \leftarrow  cpr \times \langle(\hat{\boldsymbol{\theta}}_n, \boldsymbol{u} \rangle - \beta ||\boldsymbol{u}||_{\boldsymbol{\Lambda}_n^{-1}}) + (1-cpr) \times f^B_{\boldsymbol{W}^{(t-1)}}, \forall \boldsymbol{u} \in \mathbb{R}^d$\;
\KwOut{$\hat{\pi}(\boldsymbol{x}) \leftarrow \arg \max_{a \in [K]} L(\boldsymbol{x}_a),  C_j^{\phi, \alpha}(\boldsymbol{u})|_{j = n+1}^{n+n_1}$}
\end{algorithm}

\subsection{Neural Sepsis Prediction with Conformal Prediction and LinLCB (NeuroSep-CP-LinLCB)}

While the NeuroSep-CP-LCB algorithm demonstrates significant potential for early sepsis prediction in ICU patients, the complexity of its neural network architecture introduces challenges related to computational efficiency and convergence. To address these limitations, we propose NeuroSep-CP-LinLCB, a secondary algorithm that extends NeuroSep-CP-LCB by leveraging a traditional contextual bandit framework with linear models. This approach allows for faster convergence and improved interpretability while maintaining the reliability of uncertainty quantification through conformal prediction intervals. 


The NeuroSep-CP-LCB framework inspires the NeuroSep-CP-LinLCB algorithm but replaces the neural network model with a more straightforward linear model for reward prediction. Given the same sepsis dataset used for NeuroSep-CP-LCB, which consists of clinical features, actions, and observed rewards, NeuroSep-CP-LinLCB is designed to optimize decision-making between septic (class 1) and non-septic (class 0) states. The algorithm integrates linear confidence bounds with conformal prediction to provide interpretable and robust decision-making capabilities. 

The algorithm initializes the confidence matrix $\boldsymbol{\Lambda}_0$ as a regularized identity matrix:
\begin{equation}
\boldsymbol{\Lambda}_0 \gets \lambda \boldsymbol{I},
\end{equation}
where $\lambda > 0$ is a regularization parameter. This ensures numerical stability during matrix inversion. The feature-action covariance matrix $\boldsymbol{\Lambda}_n$ is then updated using the offline training dataset $\mathcal{D}_n = \{(\boldsymbol{x}_t, a_t, r_t)\}_{t=1}^n$:
\begin{equation}
\boldsymbol{\Lambda}_n \leftarrow \lambda \boldsymbol{I} + \sum_{t=1}^{n} \boldsymbol{x}_{t,a_t} \boldsymbol{x}_{t,a_t}^T.
\end{equation}
The reward prediction parameters $\hat{\boldsymbol{\theta}}_n$ are computed as:
\begin{equation}
\hat{\boldsymbol{\theta}}_n \leftarrow \boldsymbol{\Lambda}_n^{-1} \sum_{t=1}^{n} \boldsymbol{x}_{t,a_t} r_t,
\end{equation}
where $\boldsymbol{x}_{t,a_t}$ represents the clinical features for the chosen action $a_t$, and $r_t$ is the corresponding reward.


Following the same way in NeuroSep-CP-LCB, NeuroSep-CP-LinLCB integrates conformal prediction intervals by invoking the subroutine \texttt{EnsembleCP1}, which computes leave-one-out (LOO) estimators and prediction intervals.

For each context-action pair $\boldsymbol{u} \in \mathbb{R}^d$, the algorithm computes a combined confidence-bound score $L(\boldsymbol{u})$ that balances the linear model predictions and neural network outputs:
\begin{equation}
L(\boldsymbol{u}) \leftarrow cpr \cdot \langle\hat{\boldsymbol{\theta}}_n, \boldsymbol{u} \rangle - \beta \|\boldsymbol{u}\|_{\boldsymbol{\Lambda}_n^{-1}} + (1-cpr) \cdot f^B_{\boldsymbol{W}^{(t-1)}}(\boldsymbol{u}),
\end{equation}
where:
\begin{itemize}
    \item $cpr \in [0,1]$ is the conformal prediction parameter that adjusts the balance between linear model outputs and neural predictions,
    \item $\beta > 0$ is the confidence scaling parameter controlling the exploration-exploitation trade-off,
    \item $\|\boldsymbol{u}\|_{\boldsymbol{\Lambda}_n^{-1}}$ represents the uncertainty of the linear model for input $\boldsymbol{u}$.
\end{itemize}

The action $\hat{\pi}(\boldsymbol{x})$ that maximizes the confidence-bound score is selected as:
\begin{equation}
\hat{\pi}(\boldsymbol{x}) \leftarrow \arg \max_{a \in [K]} L(\boldsymbol{x}_a),
\end{equation}
where $[K]$ is the set of possible actions (e.g., septic or nonseptic).
The algorithm outputs both the selected action $\hat{\pi}(\boldsymbol{x})$ and the conformal prediction intervals $C_j^{\phi, \alpha}(\boldsymbol{u})$ for new data points $j = n+1, \dots, n+n_1$.



The proposed NeuroSep-CP-LinLCB algorithm extends the NeuroSep-CP-LCB framework by integrating traditional contextual bandit methods with conformal prediction. This adaptation demonstrates the NeuroSep framework's flexibility and applicability to various modeling approaches, making it a promising solution for early sepsis detection in ICU patients.

The motivation for proposing NeuroSep-CP-LinLCB lies in the inherent characteristics of the dataset and the computational challenges posed by the primary algorithm (NeuroSep-CP-LCB):

\begin{itemize}
    \item \textbf{Dataset Temporal Dependencies Characteristics:} 
The sepsis dataset used in this work consists of clinical data collected at regular time intervals for ICU patients, making it inherently time-series.
Time-series data often exhibits sequential patterns, where the evolution of features over time is crucial for making predictions. Linear models are well-suited to leverage such sequential trends, especially when the data is not overly complex or non-linear.
In practice, simpler models like linear contextual bandits (e.g., NeuroSep-CP-LinLCB) can efficiently capture short-term trends in the data without overfitting or requiring extensive parameter tuning.
\item \textbf{Interpretability and Simplicity:}
Healthcare applications often prioritize interpretability for clinicians. Linear models provide transparent and interpretable decision boundaries compared to neural networks, often considered "black-box" models.
A linear approach allows for clearer explanations of how each clinical feature (e.g., heart rate, blood pressure) influences the decision. This is critical for building trust in automated decision-making systems for high-stakes tasks like sepsis prediction.
\item \textbf{Faster Convergence and Computational Efficiency:} Neural network-based algorithms, such as NeuroSep-CP-LCB, often require larger datasets and extensive training to achieve reliable performance. In contrast, linear models like NeuroSep-CP-LinLCB are computationally efficient and converge faster, even with smaller datasets or limited computational resources.
In a real-time ICU setting, where predictions must be made quickly as new patient data arrives, NeuroSep-CP-LinLCB's lightweight nature makes it more practical.
\item \textbf{Handling Overfitting in Sequential Data:}  
Neural networks tend to overfit when trained on small or noisy datasets, which can be a concern with clinical datasets that may have limited samples per patient. A linear model naturally incorporates regularization (via $\lambda$ in the confidence matrix) to mitigate overfitting.
For sequential, time-series-like data, where changes in feature values are gradual, a linear model can effectively generalize without introducing unnecessary complexity.
\item \textbf{Balancing Flexibility with Robustness:} 
While NeuroSep-CP-LCB provides flexibility in capturing non-linear patterns, its performance may be sensitive to hyperparameter choices such as learning rate ($lr$) and confidence scaling ($\beta$). Linear contextual bandits are more robust to such variations, making them a reliable alternative in scenarios with time-varying patterns or less predictable behavior.
\end{itemize}

\section{Results}
\label{sec:res}

In this section, we evaluate the performance of NeuroSep-CP-LCB and NeuroSep-CP-LinLCB  by their generalization ability using the expected sub-optimality.
For each algorithm, we vary the number of training samples $n$ from 1 to $T$ (from 10,000 to 27,000).\footnote{Due to the memory limitation, we have different $T$ for different hyperparameters. NeuroSep-CP-LCB demands more memory than NeuroSep-CP-LinLCB, so NeuroSep-CP-LinLCB plots usually have a larger $T$.}
For data imputation, we employed the \textit{\textbf{HyperImpute}} \cite{jarrett2022hyperimpute}, a generalized iterative imputation framework that can adaptively and automatically configure column-wise models and the corresponding hyperparameters, to efficiently handle missing data and ensure robust and adaptive imputation performance. And we split each patient's data into an eight-hour window. We have a balanced training dataset with up to 1,687 septic patients' windows and 1,687 non-septic patients' windows. We have 250 septic patients' windows and 3,000 non-septic patients' windows in the testing dataset. 

To enhance computational efficiency, we adopt the same strategies as prior works \cite{riquelme2018deep,zhou2020neural,nguyen-tang2022offline} to simplify the computation of large covariance matrices and resource-intensive kernel methods. Specifically, both NeuroSep-CP-LCB and NeuroSep-CP-LinLCB require the computation of the covariance matrix $\boldsymbol{\Lambda}_t$ of size $p \times p$, where $p = md + m + m^2(L-2)$ represents the total number of neural network parameters. We assigned the layer size $m$ as 100 for all the experiments. Given the large dimensionality, we approximate $\boldsymbol{\Lambda}_t$ by retaining only its diagonal elements.  

Offline actions in the training dataset are generated using an $\epsilon$-greedy policy aligned with the true reward function. The value of $\epsilon$ is set to 0.1 for all experimental runs. In each run with different train sizes, we randomly sample 250 septic patients' windows and 3,000 non-septic patients' windows (i.e., 26,000 hours of records) to estimate the expected sub-optimality of each algorithm's actions. We conducted experiments exclusively on the septic group to showcase the performance of the positive group, as in the context of early sepsis prediction, the performance is especially critical when dealing with actual septic patients. For the conformal prediction hyperparameters, we assign the number of bootstrap models $B=10$ for all the experiments.

\begin{table}[!ht]
\centering
\caption{Average Regret of NeuroSep-CP-LCB with Different Hyperparameters.}
\begin{tabular}{|c|c|c|c|c|c|}
\hline
\hline
\textbf{Algorithm} & \textbf{lr} & $\beta$ & \textbf{Group} & \textbf{Train Size} & \textbf{Regret} \\
\hline
NeuroSep-CP-LCB & 0.0001 & 1.0 & septic \& nonseptic & 1 & 0.1326 \\
NeuroSep-CP-LCB & 0.0001 & 1.0 & septic \& nonseptic & 2001 & 0.0108 \\
NeuroSep-CP-LCB & 0.0001 & 1.0 & septic \& nonseptic & 4001 & 0.0250 \\
NeuroSep-CP-LCB & 0.0001 & 1.0 & septic \& nonseptic & 6001 & 0.0196 \\
NeuroSep-CP-LCB & 0.0001 & 1.0 & septic \& nonseptic & 8001 & 0.0224 \\
NeuroSep-CP-LCB & 0.0001 & 5.0 & septic \& nonseptic & 1 & 0.3208 \\
NeuroSep-CP-LCB & 0.0001 & 5.0 & septic \& nonseptic & 2001 & 0.0110 \\
NeuroSep-CP-LCB & 0.0001 & 5.0 & septic \& nonseptic & 4001 & 0.0370 \\
NeuroSep-CP-LCB & 0.0001 & 5.0 & septic \& nonseptic & 6001 & 0.0150 \\
NeuroSep-CP-LCB & 0.0001 & 5.0 & septic \& nonseptic & 8001 & 0.0190 \\
\hline
\hline
\end{tabular}
\label{tab:ApproxNeuraLCB_cp1}
\end{table}

For implication details, please refer to our open-sourced code.\footnote{\url{https://github.com/Annie983284450-1/NeuralLCB_C.git}} All computational experiments were performed using the resources provided by the Partnership for an Advanced Computing Environment (PACE) \cite{PACE} at the Georgia Institute of Technology.  

\subsection{Regret Analysis of NeuroSep-CP-LCB}

\begin{figure}[!ht]
    \centering
    \includegraphics[width=\linewidth]{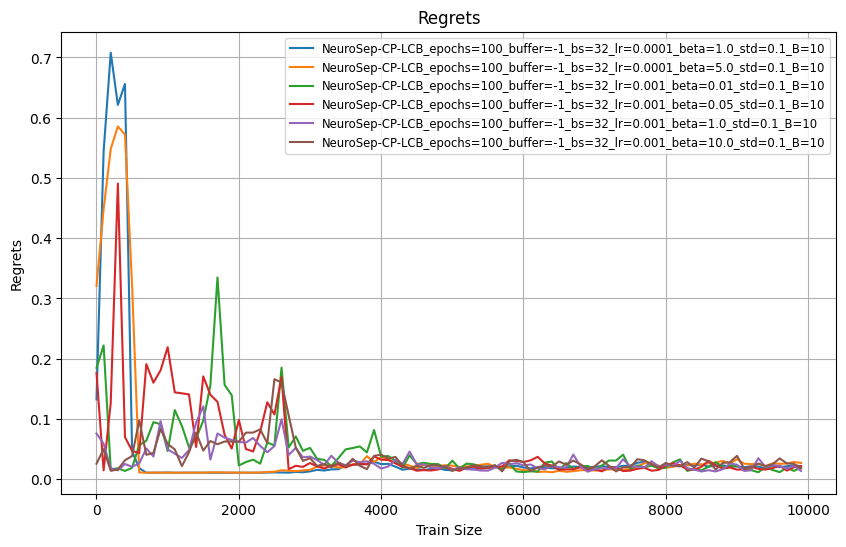}
    \caption{Average Regret of NeuroSep-CP-LCB with different $\beta$ values and learning rates for mixed dataset.}
    \label{fig:neuraLCBall}
\end{figure}

The results in \ref{tab:ApproxNeuraLCB_cp1}, \ref{tab:ApproxNeuraLCB_cp2}, \ref{tab:ApproxNeuraLCB_cpG1}, \ref{fig:neuraLCBall}, and \ref{fig:neuraLCBLr1e3G1} highlight the algorithm's behavior under different hyperparameter settings, focusing on confidence scaling parameter ($\beta$), learning rate ($lr$), and group-specific performance. Below, we comprehensively analyze the observations and possible reasons for the observed trends.
\begin{table}[!ht]
\centering
\caption{Average Regret of NeuroSep-CP-LCB with Different Hyperparameters.}
\begin{tabular}{|c|c|c|c|c|c|}
\hline
\hline
\textbf{Algorithm} & \textbf{lr} & $\beta$ & \textbf{Group} & \textbf{Train Size} & \textbf{Regret} \\
\hline
NeuroSep-CP-LCB & 0.001 & 0.01 & septic \& nonseptic & 1 & 0.1848 \\
NeuroSep-CP-LCB & 0.001 & 0.01 & septic \& nonseptic & 2001 & 0.0229 \\
NeuroSep-CP-LCB & 0.001 & 0.01 & septic \& nonseptic & 4001 & 0.0376 \\
NeuroSep-CP-LCB & 0.001 & 0.01 & septic \& nonseptic & 6001 & 0.0119 \\
NeuroSep-CP-LCB & 0.001 & 0.01 & septic \& nonseptic & 8001 & 0.0192 \\
NeuroSep-CP-LCB & 0.001 & 0.05 & septic \& nonseptic & 1 & 0.1765 \\
NeuroSep-CP-LCB & 0.001 & 0.05 & septic \& nonseptic & 2001 & 0.0983 \\
NeuroSep-CP-LCB & 0.001 & 0.05 & septic \& nonseptic & 4001 & 0.0324 \\
NeuroSep-CP-LCB & 0.001 & 0.05 & septic \& nonseptic & 6001 & 0.0287 \\
NeuroSep-CP-LCB & 0.001 & 0.05 & septic \& nonseptic & 8001 & 0.0231 \\
\hline
\hline
\end{tabular}
\label{tab:ApproxNeuraLCB_cp2}
\end{table}


\ref{tab:ApproxNeuraLCB_cp1}, \ref{tab:ApproxNeuraLCB_cp2}, \ref{tab:ApproxNeuraLCB_cpG1}, \ref{fig:neuraLCBall}, and \ref{fig:neuraLCBLr1e3G1} show that the average regret decreases as the training size increases, demonstrating the NeuroSep-CP-LCB's ability to learn and adapt as more data becomes available:
\begin{itemize}
    \item In \ref{tab:ApproxNeuraLCB_cp1}, for $lr=0.0001$ and $\beta=1.0$, the regret decreases from $0.1326$ when train size is $1$ to $0.0224$ when train size is $8,001$.
    \item Similarly, in \ref{fig:neuraLCBall}, the average regrets for all configurations of NeuroSep-CP-LCB stabilize and converge to near-zero values as the training size exceeds 2,000 samples.
\end{itemize}

\begin{table}[!ht]
\centering
\caption{Average Regret (Group: septic, Batch Size = 32).}
\begin{tabular}{|c|c|c|c|c|c|}
\hline
\hline
\textbf{Algorithm} & $lr$ & $\beta$ & \textbf{Group} & \textbf{Train Size} & \textbf{Regret} \\
\hline
NeuroSep-CP-LCB & 0.0001 & 0.5 & septic & 1 & 0.1580 \\
NeuroSep-CP-LCB & 0.0001 & 0.5 & septic & 2001 & 0.1420 \\
NeuroSep-CP-LCB & 0.0001 & 0.5 & septic & 4001 & 0.1420 \\
NeuroSep-CP-LCB & 0.0001 & 0.5 & septic & 6001 & 0.1485 \\
NeuroSep-CP-LCB & 0.0001 & 0.5 & septic & 8001 & 0.1445 \\
NeuroSep-CP-LCB & 0.0001 & 1.0 & septic & 1 & 0.1605 \\
NeuroSep-CP-LCB & 0.0001 & 1.0 & septic & 2001 & 0.1415 \\
NeuroSep-CP-LCB & 0.0001 & 1.0 & septic & 4001 & 0.1520 \\
NeuroSep-CP-LCB & 0.0001 & 1.0 & septic & 6001 & 0.1565 \\
NeuroSep-CP-LCB & 0.0001 & 1.0 & septic & 8001 & 0.1690 \\
NeuroSep-CP-LCB & 0.0001 & 5.0 & septic & 1 & 0.3680 \\
NeuroSep-CP-LCB & 0.0001 & 5.0 & septic & 2001 & 0.1410 \\
NeuroSep-CP-LCB & 0.0001 & 5.0 & septic & 4001 & 0.1520 \\
NeuroSep-CP-LCB & 0.0001 & 5.0 & septic & 6001 & 0.1430 \\
NeuroSep-CP-LCB & 0.0001 & 5.0 & septic & 8001 & 0.1440 \\
\hline
\hline
\end{tabular}
\label{tab:ApproxNeuraLCB_cpG1}
\end{table}

\subsubsection{Effect of $\beta$}
\begin{figure}[!ht]
    \centering
    \includegraphics[width=\linewidth]{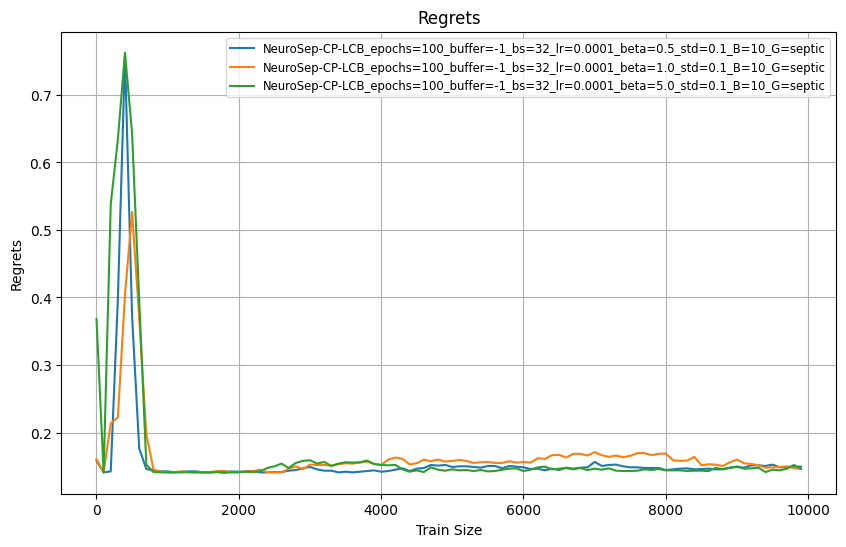}
    \caption{Average regret of NeuroSep-CP-LCB with different $\beta$ values for septic patients.}
    \label{fig:neuraLCBLr1e3G1}
\end{figure}

\ref{fig:neuraLCBall} and \ref{fig:neuraLCBLr1e3G1} show significant fluctuations in average regrets during the initial stages of training when the training size is small.
When the testing set contains a mixture of septic and non-septic patients (see \ref{fig:neuraLCBall}), configurations with higher $\beta$ values (e.g., $\beta=5.0, lr = 0.001$) experience more minor fluctuations compared to lower $\beta$ values (e.g., $\beta=0.05, lr = 0.001$). This can be attributed to increased exploration induced by higher confidence scaling parameters.
 
\paragraph{Effect of $\beta$ in Lower Confidence Bound Calculation:}
Recall that the confidence parameter $\beta$ directly affects the computation of the lower confidence bound (LCB) as shown in \ref{alg:neuralcb_bmode} (Line 4):
\[
    L_t(\boldsymbol{u}) \gets f^B_{\boldsymbol{W}^{(t-1)}}(\boldsymbol{u}) - \beta_t \| \nabla f_{\boldsymbol{W}^{(t-1)}}(\boldsymbol{u}) \|_{\boldsymbol{\Lambda}_{t-1}^{-1}}, \forall \boldsymbol{u} \in \mathbb{R}^d,
\]
in which $f^B_{\boldsymbol{W}^{(t-1)}}(\boldsymbol{u})$ is the predicted reward using the neural network and $\beta_t \| \nabla f_{\boldsymbol{W}^{(t-1)}}(\boldsymbol{u}) \|_{\boldsymbol{\Lambda}_{t-1}^{-1}}$ adjusts the confidence interval for exploration.
Larger values of $\beta$ lead to wider confidence intervals, promoting more extensive exploration of actions during the early stages of training. This might reduce the likelihood of the algorithm prematurely settling on suboptimal actions, which can happen with smaller $\beta$ values where the confidence intervals are narrower.
In addition, the policy $\pi_t(x)$ selects actions by maximizing $L_t(\boldsymbol{x}_t, a_t)$ over all possible actions $a$:
\[
\hat{\pi}_t(\boldsymbol{x}) \gets \arg\max_{a \in [K]} L_t(\boldsymbol{x}_t, a_t), \forall  \boldsymbol{x}_t \in \mathbb{R}^d , a_t \in [K].
\]
With larger $\beta$, the LCB incorporates more uncertainty into the decision-making process, effectively encouraging the exploration of less certain actions. This might result in better coverage of the action space during training and reduce regret in the long run.

The update of the confidence matrix $\Lambda_t$ depends on the gradients of the model's output:
\[
\boldsymbol{\Lambda}_t \gets \boldsymbol{\Lambda}_{t-1} + \frac{\mathrm{vec}(\nabla f_{\boldsymbol{W}^{(t-1)}}(\boldsymbol{x}_t, a_t)) \cdot \mathrm{vec}(\nabla f_{\boldsymbol{W}^{(t-1)}}(\boldsymbol{x}_t, a_t))^T}{m},
\]
By encouraging broader exploration, larger $\beta$ values ensure that the updates to $\Lambda_t$ reflect a more diverse set of state-action pairs. This results in a more robust and accurate representation of the uncertainty over time.

As shown in \ref{fig:neuraLCBall}, the initial fluctuations observed in the regret curves for mixed dataset are mitigated with larger $\beta$ because the wider confidence intervals allow the algorithm to explore more actions effectively. This helps the model avoid overfitting to the limited initial training data and reduces the impact of early suboptimal decisions.

NeuroSep-CP-LCB updates include a regularization term:
\[
        \mathcal{L}^{(j)}_t(\boldsymbol{W}) \gets \frac{1}{2B} \sum_{q=1}^B (f_{\boldsymbol{W}}(\boldsymbol{x}_{t_q}, a_{t_q}) - r_{t_q})^2 + \frac{m \lambda}{2} \|\boldsymbol{W} - \boldsymbol{W}^{(0)}\|_F^2.
\]
While this regularization ensures stability in model updates, the choice of $\beta$ affects the gradients used in the optimization. Larger $\beta$ values might indirectly lead to smoother updates by focusing on a broader exploration of the action space, reducing the sensitivity to noise in the training data.

\subsubsection{NeuroSep-CP-LCB Regret Analysis for Septic-Only Data}
The regret behavior of NeuroSep-CP-LCB on the septic-only dataset, as shown in Figure~\ref{fig:neuraLCBLr1e3G1}, reveals that the choice of $\beta$ significantly influences the algorithm's performance in the initial stages. Specifically, $\beta = 1$ yields the lowest regrets during the early training iterations, outperforming smaller ($\beta = 0.5$) and larger ($\beta = 5.0$) values of $\beta$. This observation contrasts with the results obtained for the mixed dataset containing both septic and non-septic patients, where larger values of $\beta$ generally performed better across all stages. 
 
The septic-only dataset represents a more homogeneous group of contexts and outcomes, as all data points are related to septic patients. In a septic-only dataset, the variability in the reward function is reduced compared to the mixed dataset, where septic and non-septic patient groups might have divergent reward structures. In addition, homogeneity allows for faster convergence when the exploration-exploitation balance is optimized, as the algorithm does not need to generalize across fundamentally different patient groups.

For septic-only data, $\beta = 1$ (see \ref{tab:ApproxNeuraLCB_cpG1}) balances exploration and exploitation during the early training phase. A smaller $\beta$ (e.g., $\beta = 0.5$) leads to narrower confidence intervals, which restricts exploration and increases the likelihood of early suboptimal decisions, as seen in the higher initial regrets. A larger $\beta$ (e.g., $\beta = 5.0$) promotes extensive exploration, which may be unnecessary in the septic-only dataset where the action space and reward variability are more constrained.
As a result, the intermediate $\beta = 1$ effectively leverages the lower variability in the septic-only dataset, achieving more efficient learning and faster convergence during the initial stages.
In contrast, the mixed dataset requires larger $\beta$ values to navigate the increased uncertainty arising from the heterogeneity of the patient population (see \ref{tab:ApproxNeuraLCB_cp1}).


The consistently higher regret in the septic group, which is normal in early sepsis prediction problem, is likely due to:
\begin{itemize}
    \item \textbf{Smaller Sample Size:} The septic group represents a subset of the data, limiting the algorithm's ability to generalize and optimize its decisions.
    \item \textbf{Higher Complexity:} Sepsis cases often involve more complex and variable clinical features, making it harder for the algorithm to learn an effective policy than the mixture of septic and non-septic patient datasets, which benefit from more diverse and larger data.
\end{itemize}

The superior performance of $\beta = 1$ in the septic-only dataset underscores the importance of tuning $\beta$ to match the dataset's characteristics. While larger $\beta$ values are advantageous in diverse and heterogeneous datasets, intermediate values are optimal for homogeneous datasets like septic-only data, where variability is reduced. This result highlights the adaptability of NeuroSep-CP-LCB to different clinical scenarios, emphasizing the need for dataset-specific parameter tuning.

In conclusion, the regret results for NeuroSep-CP-LCB demonstrate its ability to adapt and learn effective decision-making policies, even for high-stakes tasks like early sepsis prediction. While the septic group presents unique challenges with higher initial regret and fluctuations, the algorithm converges as expected, achieving low regret across all configurations and validating the robustness of the NeuroSep-CP-LCB algorithm. The choice of $\beta$ is critical in balancing exploration and exploitation, particularly for smaller datasets or groups with higher variability.
\begin{figure}[!ht]
\centering
\includegraphics[width=\textwidth]{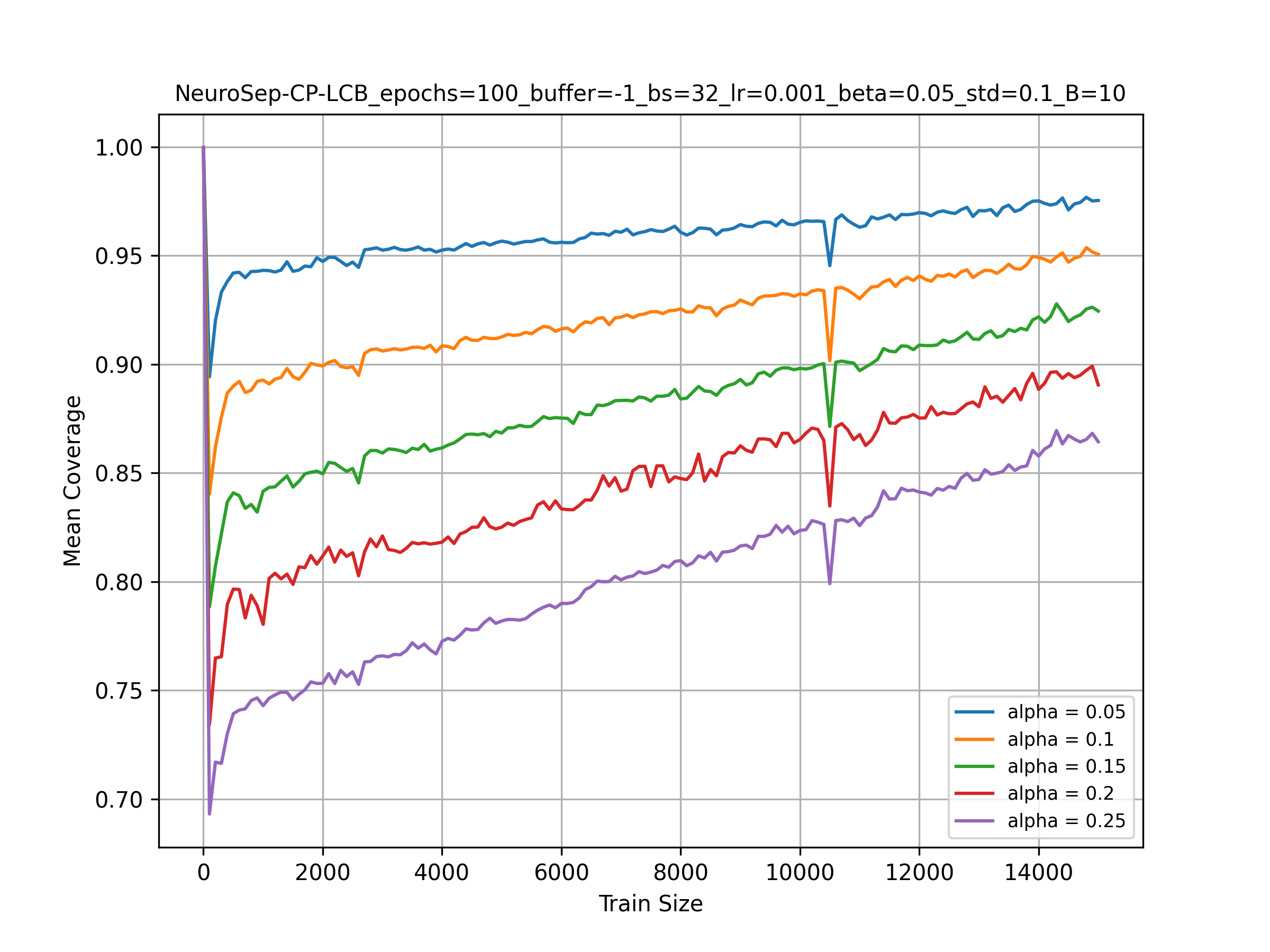}
\caption{Mean Coverage for NeuroSep-CP-LCB with lr=0.001, $\beta=0.05$.}
\label{fig:ApproxNeuraLCB_cp_epochs100_buffer-1_bs32_lr0_001_beta0_05_std0_1_B10_PIs_Coverage}
\end{figure}

\begin{figure}[ht]
\centering
\includegraphics[width=\textwidth]{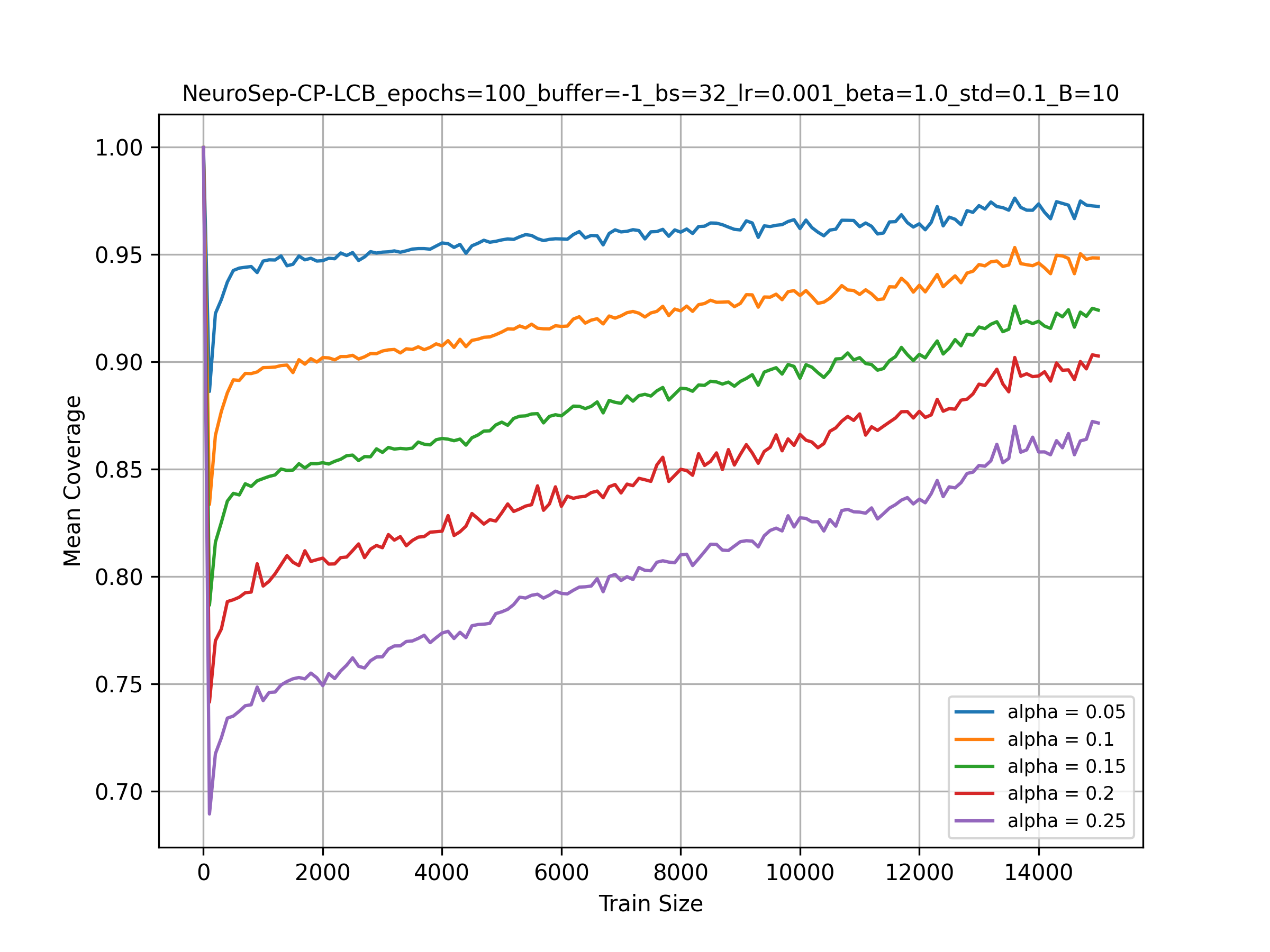}
\caption{Mean Coverage for NeuroSep-CP-LCB with lr=0.001, $\beta=1.0$.}
\label{fig:ApproxNeuraLCB_cp_epochs100_buffer-1_bs32_lr0_001_beta1_0_std0_1_B10_PIs_Coverage}
\end{figure}

\subsection{Prediction Intervals Analysis for NeuroSep-CP-LCB}
In this section, we demonstrate the prediction interval (PI) performance of the NeuroSep-CP-LCB algorithm under various configurations of learning rates $lr$, confidence scaling parameters $\beta$, and significance levels $\alpha$. 
The analysis includes two key metrics: \textit{\textbf{mean coverage}} and \textit{\textbf{average width}} of prediction intervals.  
\subsubsection{Mean Coverage of Prediction Intervals}
The mean coverage figures depict how well the conformal prediction intervals capture the true outcomes across training sizes and parameter settings.

\paragraph{Effect of Conformal Prediction Significance Level $\alpha$:}
The results (e.g., \ref{fig:ApproxNeuraLCB_cp_epochs100_buffer-1_bs32_lr0_001_beta0_05_std0_1_B10_PIs_Coverage}, \ref{fig:ApproxNeuraLCB_cp_epochs100_buffer-1_bs32_lr0_001_beta1_0_std0_1_B10_PIs_Coverage}) demonstrate a clear relationship between the significance level $\alpha$ and the mean coverage:
\begin{itemize}
    \item For smaller values of $\alpha$ (e.g., $\alpha=0.05$), the mean coverage approaches or exceeds 95\%, as expected from the definition of conformal prediction.
    \item As $\alpha$ increases (e.g., $\alpha=0.25$), the mean coverage decreases steadily, reflecting narrower prediction intervals that are less likely to cover the true values.
\end{itemize}
For more comprehensive experimental results on the effect of $\alpha$ of different configurations, please refer to \ref{sec:appendix_neuralcb}.

\paragraph{Effect of Confidence Parameter $\beta$:}
Comparing the configuration of $\beta=0.05$ (\ref{fig:ApproxNeuraLCB_cp_epochs100_buffer-1_bs32_lr0_001_beta0_05_std0_1_B10_PIs_Coverage}) and $\beta=1.0$ (\ref{fig:ApproxNeuraLCB_cp_epochs100_buffer-1_bs32_lr0_001_beta1_0_std0_1_B10_PIs_Coverage}) reveals that smaller $\beta$ values (e.g., $\beta=0.05$, \ref{fig:ApproxNeuraLCB_cp_epochs100_buffer-1_bs32_lr0_001_beta0_05_std0_1_B10_PIs_Coverage}) show  more stable coverage as training progresses. 


\subsubsection{Average Width of Prediction Intervals}

The average width of PIs reflects the uncertainty associated with predictions. Narrower intervals indicate more confident predictions, while wider intervals suggest higher uncertainty.

\paragraph{Effect of Conformal Prediction Significance Level $\alpha$:}
As expected, higher $\alpha$ values lead to narrower prediction intervals, as the algorithm relaxes its coverage requirements:
\begin{itemize}
    \item For $\alpha=0.05$, the PIs are the widest, ensuring high coverage but potentially reducing practical utility due to over-conservatism.
    \item For $\alpha=0.25$, the PIs are significantly narrower, providing tighter but less reliable predictions.
\end{itemize}

\paragraph{Effect of Confidence Parameter $\beta$:}
The confidence scaling parameter $\beta$ strongly influences the average width:
\begin{itemize}
    \item Higher $\beta$ values (e.g., $\beta=5.0$) result in wider intervals, reflecting increased exploration and higher uncertainty estimates.
    \item Smaller $\beta$ values (e.g., $\beta=0.05$) produce narrower intervals, indicating greater confidence in predictions.
\end{itemize}



\paragraph{Coverage-Width Trade-off:}
The results highlight the inherent trade-off between mean coverage and average width:
\begin{itemize}
    \item Higher $\alpha$ values and smaller $\beta$ values lead to narrower intervals at the cost of reduced coverage.
    \item Lower $\alpha$ values and larger $\beta$ values ensure higher coverage but produce wider intervals, which may not always be practical for decision-making.
\end{itemize}
The choice of $\alpha$ and $\beta$ plays a crucial role in balancing this trade-off. For early sepsis prediction in ICU patients, the optimal configuration depends on the desired confidence level and the application-specific tolerance for false positives versus false negatives.


The results demonstrate the flexibility of NeuroSep-CP-LCB in generating prediction intervals with tunable properties. By tuning $\alpha$ and $\beta$, the algorithm can provide reliable uncertainty estimates tailored to the needs of early sepsis prediction. The results emphasize the importance of hyperparameter tuning to achieve the desired balance between prediction reliability and practical utility in clinical settings.
\begin{figure}[!ht]
    \centering
    \includegraphics[width=\linewidth]{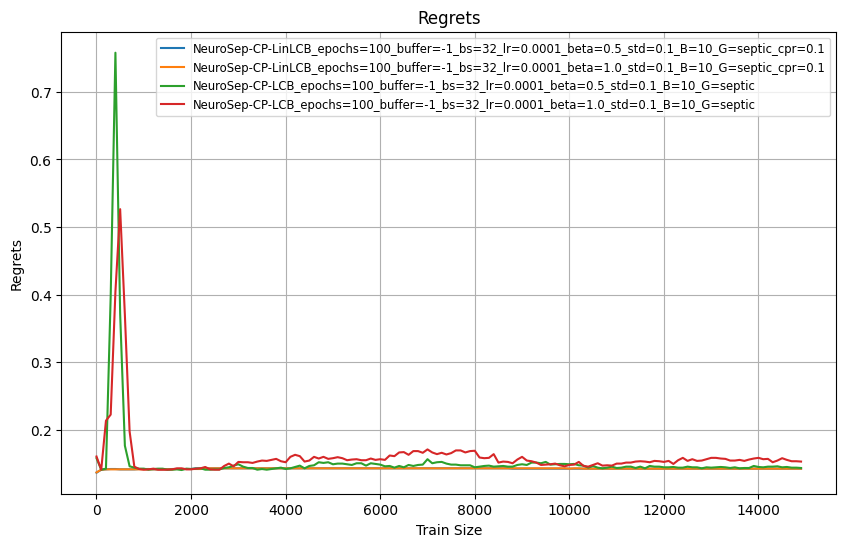}
    \caption{Average Regret of NeuroSep-CP-LCB and NeuroSep-CP-LinLCB.}
    \label{fig:neuralcbVSlin}
\end{figure}

\section{Discussion}
\label{sec:disc}
 
\subsection{Analysis of NeuroSep-CP-LinLCB Results}
\subsubsection{Comparison with NeuroSep-CP-LCB}
NeuroSep-CP-LinLCB extends the framework of NeuroSep-CP-LCB by incorporating linear contextual bandit estimators alongside the conformal prediction component. The results demonstrate distinct performance improvements in regret, mean coverage, and interval width. 

\begin{table}[!ht]
\centering
\caption{Average Regret (Group: septic \& nonseptic, $cpr = 0.5$).}
\begin{tabular}{|c|c|c|c|c|c|}
\hline
\hline
\textbf{Algorithm} & $lr$ & $\beta$ & \textbf{Group} & \textbf{Train Size} & \textbf{Regret} \\
\hline
NeuroSep-CP-LinLCB & 0.0001 & 0.05 & septic \& nonseptic & 1 & 0.0175 \\
NeuroSep-CP-LinLCB & 0.0001 & 0.05 & septic \& nonseptic & 2001 & 0.0118 \\
NeuroSep-CP-LinLCB & 0.0001 & 0.05 & septic \& nonseptic & 4001 & 0.0119 \\
NeuroSep-CP-LinLCB & 0.0001 & 0.05 & septic \& nonseptic & 6001 & 0.0118 \\
NeuroSep-CP-LinLCB & 0.0001 & 0.05 & septic \& nonseptic & 8001 & 0.0119 \\
NeuroSep-CP-LinLCB & 0.0001 & 5.0 & septic \& nonseptic & 1 & 0.1090 \\
NeuroSep-CP-LinLCB & 0.0001 & 5.0 & septic \& nonseptic & 2001 & 0.0117 \\
NeuroSep-CP-LinLCB & 0.0001 & 5.0 & septic \& nonseptic & 4001 & 0.0118 \\
NeuroSep-CP-LinLCB & 0.0001 & 5.0 & septic \& nonseptic & 6001 & 0.0118 \\
NeuroSep-CP-LinLCB & 0.0001 & 5.0 & septic \& nonseptic & 8001 & 0.0119 \\
\hline
\hline
\end{tabular}
\label{tab:linucb_cpr05}
\end{table}

\paragraph{Lower Average Regret:}
As seen in  \ref{tab:linucb_cpr05} and \ref{tab:linucb_G1cpr01}, NeuroSep-CP-LinLCB achieves significantly lower average regret compared to NeuroSep-CP-LCB:
\begin{itemize}
    \item For $lr=0.0001$ and $\beta=5.0$, the regret of NeuroSep-CP-LinLCB is $0.1090$ when train size is $1$, compared to $0.3208$ for NeuroSep-CP-LCB under the same settings (see \ref{tab:ApproxNeuraLCB_cp1}).
    \item Across all train sizes, NeuroSep-CP-LinLCB maintains a relatively stable regret (e.g., $0.0119$ when train size is $8,001$, $lr =0.0001$ and $\beta=0.05$), highlighting its robustness and fast convergence compared to the fluctuations observed in NeuroSep-CP-LCB.
\end{itemize}
\begin{table}[!ht]
\centering
\caption{Average Regret (Group: septic, $cpr = 0.1$).}
\begin{tabular}{|c|c|c|c|c|c|}
\hline
\hline
\textbf{Algorithm} & $lr$ & $\beta$ & \textbf{Group} & \textbf{Train Size} & \textbf{Regret} \\
\hline
NeuroSep-CP-LinLCB & 0.0001 & 0.01 & septic & 1 & 0.1375 \\
NeuroSep-CP-LinLCB & 0.0001 & 0.01 & septic & 2001 & 0.1430 \\
NeuroSep-CP-LinLCB & 0.0001 & 0.01 & septic & 4001 & 0.1430 \\
NeuroSep-CP-LinLCB & 0.0001 & 0.01 & septic & 6001 & 0.1430 \\
NeuroSep-CP-LinLCB & 0.0001 & 0.01 & septic & 8001 & 0.1425 \\
NeuroSep-CP-LinLCB & 0.0001 & 0.5 & septic & 1 & 0.1365 \\
NeuroSep-CP-LinLCB & 0.0001 & 0.5 & septic & 2001 & 0.1420 \\
NeuroSep-CP-LinLCB & 0.0001 & 0.5 & septic & 4001 & 0.1430 \\
NeuroSep-CP-LinLCB & 0.0001 & 0.5 & septic & 6001 & 0.1430 \\
NeuroSep-CP-LinLCB & 0.0001 & 0.5 & septic & 8001 & 0.1430 \\
NeuroSep-CP-LinLCB & 0.0001 & 1.0 & septic & 1 & 0.1370 \\
NeuroSep-CP-LinLCB & 0.0001 & 1.0 & septic & 2001 & 0.1420 \\
NeuroSep-CP-LinLCB & 0.0001 & 1.0 & septic & 4001 & 0.1430 \\
NeuroSep-CP-LinLCB & 0.0001 & 1.0 & septic & 6001 & 0.1430 \\
NeuroSep-CP-LinLCB & 0.0001 & 1.0 & septic & 8001 & 0.1430 \\
\hline
\hline
\end{tabular}
\label{tab:linucb_G1cpr01}
\end{table}
\paragraph{Improved Stability Across Hyperparameters:}
NeuroSep-CP-LinLCB shows stable regret across different $\beta$ values. For instance:
\begin{itemize}
    \item When $\beta=5.0, lr = 0.0001$, the regret stabilizes at $0.0119$ with train size reaching $8,001$ (see \ref{tab:linucb_cpr05}), similar to the results at $\beta=0.05$, indicating that the linear model contributes to more consistent exploration-exploitation balancing.
    \item In contrast, NeuroSep-CP-LCB exhibits more variation in regret as $\beta$ increases, as larger $\beta$ values introduce higher exploratory behavior leading to initial fluctuations.
    \item While larger $\beta$ values still introduce some variability, NeuroSep-CP-LinLCB shows less sensitivity to this parameter compared to NeuroSep-CP-LCB, highlighting the robustness of the combined linear and neural framework.
\end{itemize}

\begin{figure}[!ht]
    \centering
    \includegraphics[width=\linewidth]{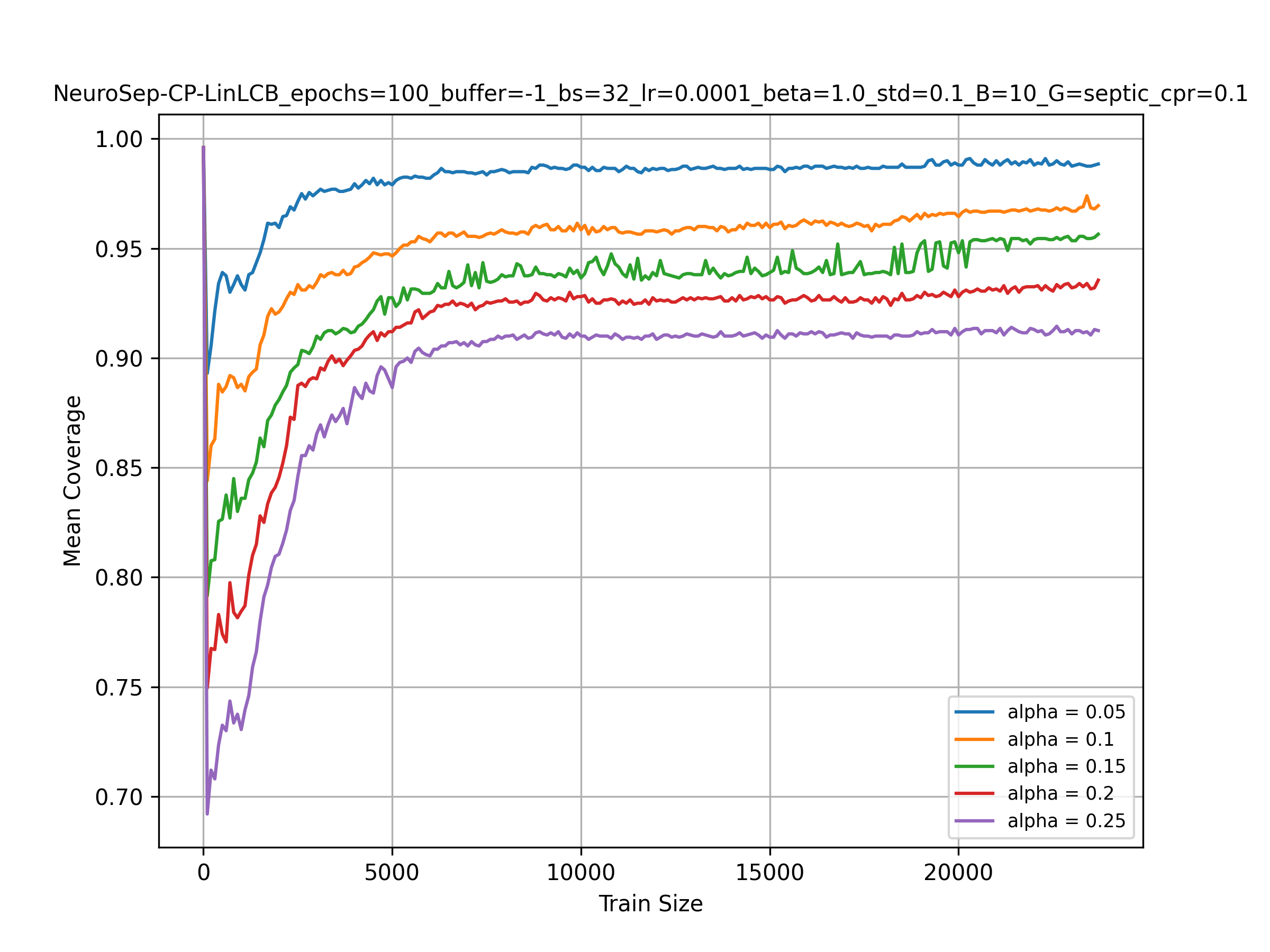}
    \caption{Mean Coverage for NeuroSep-CP-LinLCB with $lr=0.0001, \beta = 1.0$.}
    \label{fig:ApproxNeuralLinLCBJointModel_cp_epochs100_buffer-1_bs32_lr0_0001_beta1_0_std0_1_B10_G1_cpr0_1PIs_Coverage}
\end{figure}


\subsection{Prediction Interval Coverage and Width Analysis of NeuroSep-CP-LinLCB}
\subsubsection{Mean Coverage Analysis}
\begin{figure}[!ht]
\centering
\includegraphics[width=0.8\textwidth]{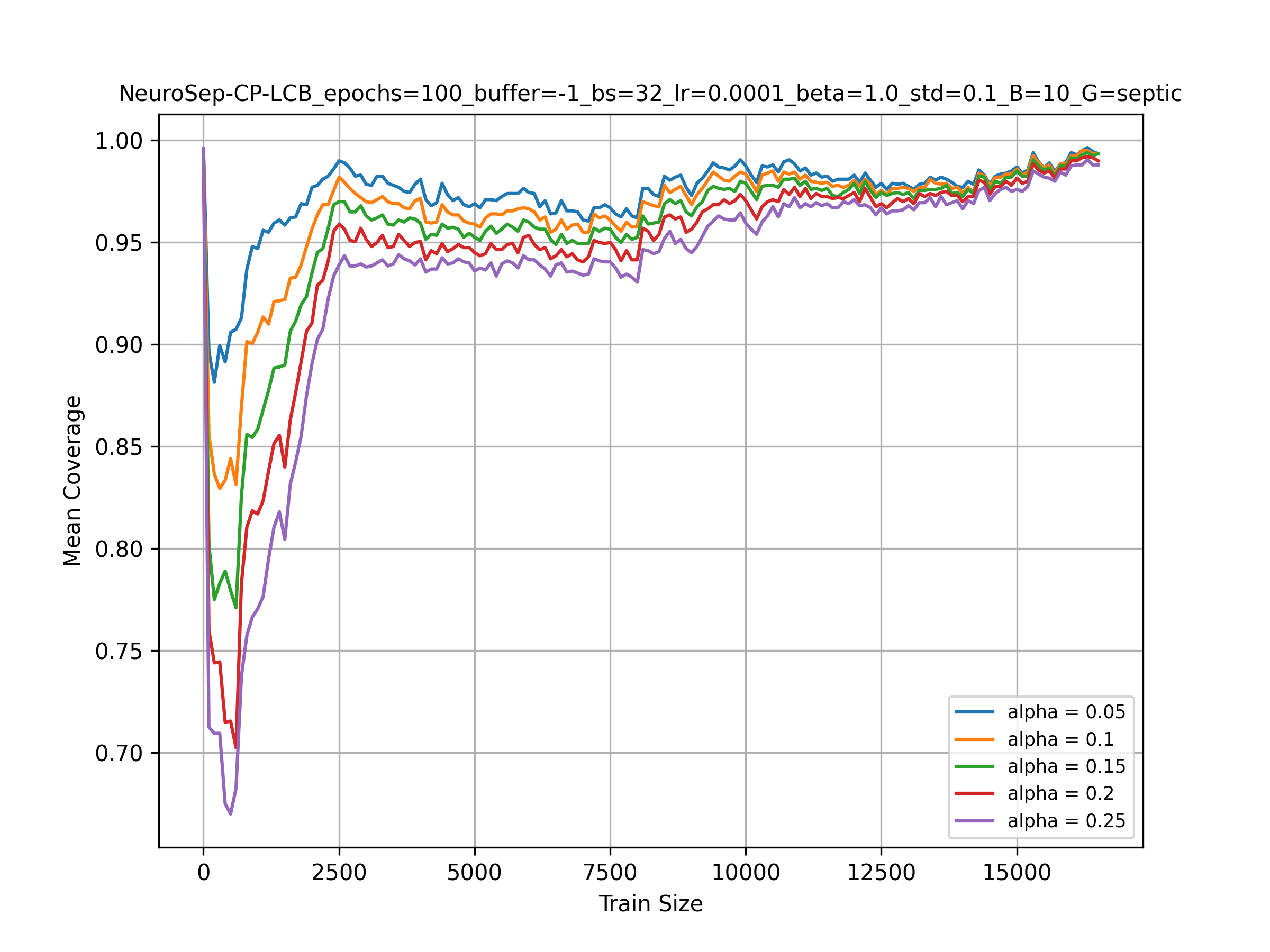}
\caption{Mean Coverage for NeuroSep-CP-LCB with lr=0.0001, $\beta=1.0$.}
\label{fig:ApproxNeuraLCB_cp_epochs100_buffer-1_bs32_lr0_0001_beta1_0_std0_1_B10_G1_PIs_Coverage}
\end{figure}

\ref{fig:ApproxNeuralLinLCBJointModel_cp_epochs100_buffer-1_bs32_lr0_0001_beta1_0_std0_1_B10_G1_cpr0_1PIs_Coverage} displays the mean coverage of prediction intervals for different $\alpha$ values:
\begin{itemize}
    \item For $\beta=0.05$, the mean coverage steadily increases across all $\alpha$ levels, converging towards the expected coverage thresholds (e.g., $95\%$ for $\alpha=0.05$).
    \item NeuroSep-CP-LinLCB demonstrates a similar pattern to NeuroSep-CP-LCB but achieves slightly higher coverage consistency, especially for larger training sizes. This improvement reflects the stability introduced by combining linear estimators with neural confidence bounds.
\end{itemize}
By comparing \ref{fig:ApproxNeuralLinLCBJointModel_cp_epochs100_buffer-1_bs32_lr0_0001_beta1_0_std0_1_B10_G1_cpr0_1PIs_Coverage} and \ref{fig:ApproxNeuraLCB_cp_epochs100_buffer-1_bs32_lr0_0001_beta1_0_std0_1_B10_G1_PIs_Coverage}, we observed that both algorithms achieve consistent coverage across different $\alpha$ values, with higher $\alpha$ values corresponding to lower coverage, as expected in conformal prediction frameworks. However, NeuroSep-CP-LinLCB demonstrates smoother convergence and slightly better coverage stability, particularly for $\alpha = 0.1$ and $\alpha = 0.05$, while NeuroSep-CP-LCB has higher coverage. Specifically, the mean coverage is larger than 95\% regardless of significance level $\alpha$.

\begin{figure}[ht]
    \centering
    \includegraphics[width=\linewidth]{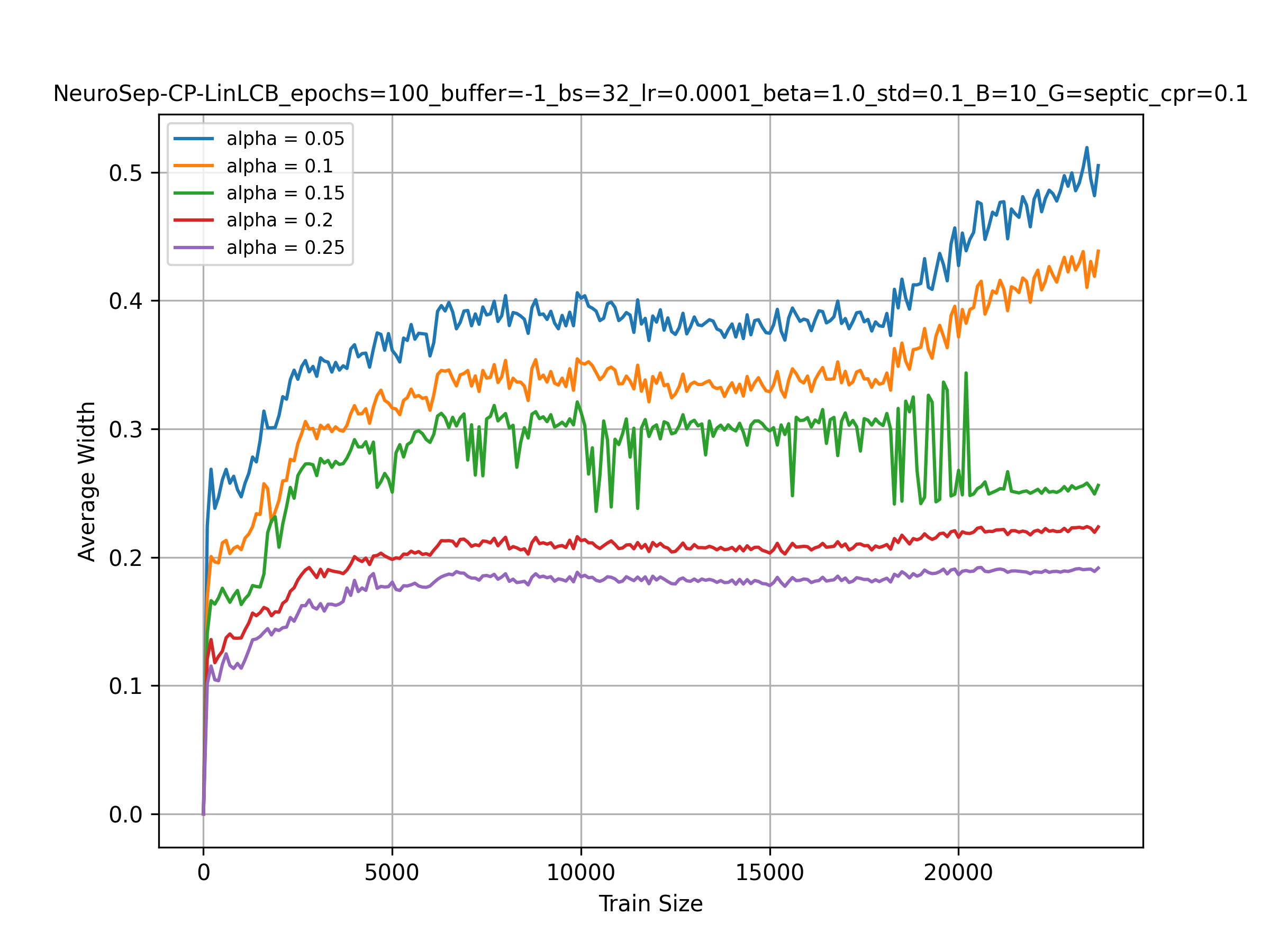}
    \caption{Average Width for NeuroSep-CP-LinLCB with $lr=0.0001, \beta = 1.0$.}
    \label{fig:ApproxNeuralLinLCBJointModel_cp_epochs100_buffer-1_bs32_lr0_0001_beta1_0_std0_1_B10_G1_cpr0_1PIs_Width}
\end{figure}

For NeuroSep-CP-LCB, initial fluctuations in coverage are more pronounced, especially for larger $\alpha$ values such as $\alpha = 0.25$. This behavior can be attributed to the neural network's reliance on sufficient training data to stabilize its confidence intervals. In contrast, the linear model in NeuroSep-CP-LinLCB benefits from the reduced complexity of the septic-only dataset, leading to faster stabilization. 

\subsubsection{Average Width Analysis}

\begin{figure}[!ht]
\centering
\includegraphics[width=0.8\textwidth]{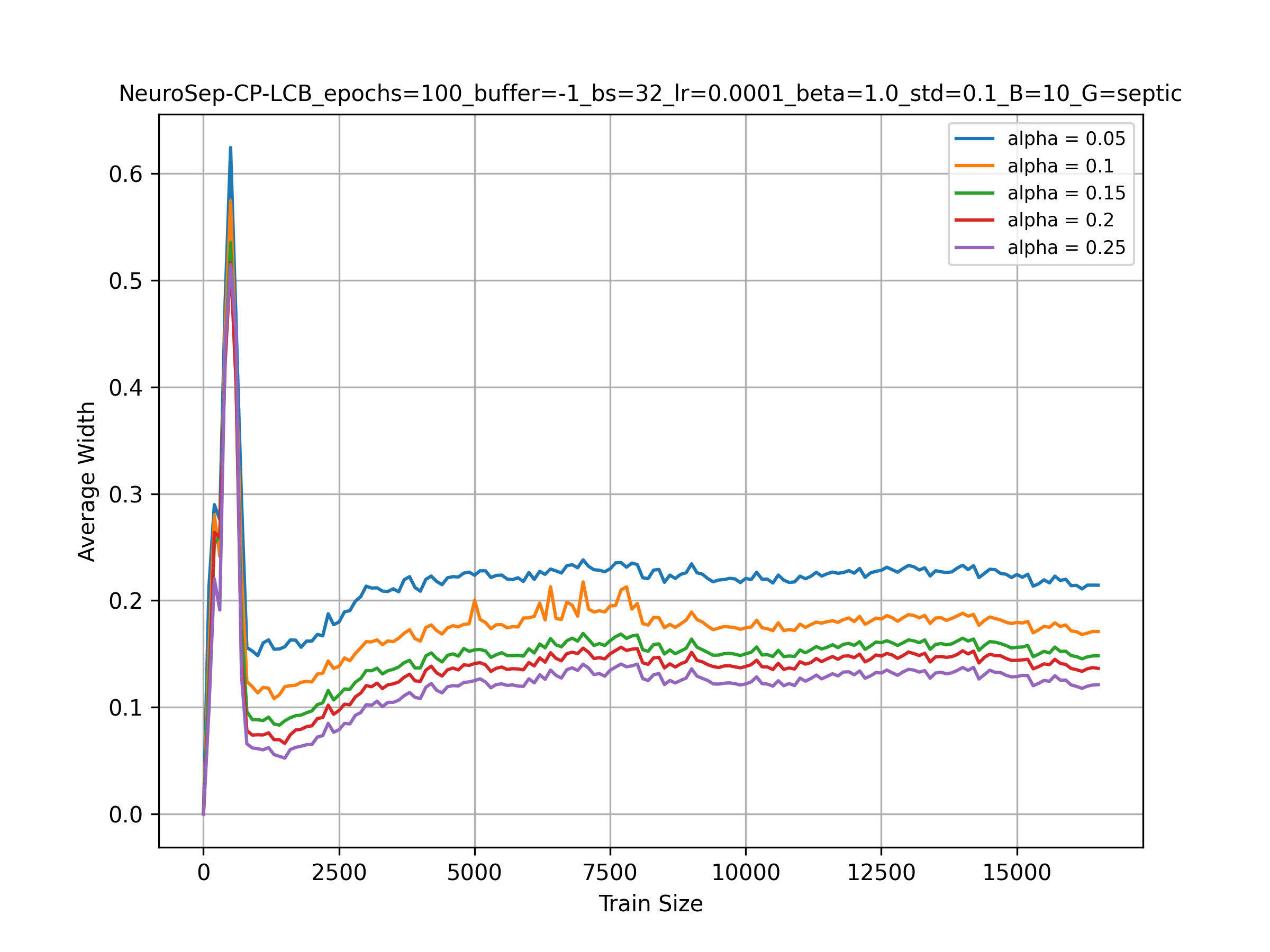}
\caption{Average Width for NeuroSep-CP-LCB with lr=0.0001, $\beta=1.0$.}
\label{fig:ApproxNeuraLCB_cp_epochs100_buffer-1_bs32_lr0_0001_beta1_0_std0_1_B10_G1_PIs_Width}
\end{figure}

\ref{fig:ApproxNeuralLinLCBJointModel_cp_epochs100_buffer-1_bs32_lr0_0001_beta1_0_std0_1_B10_G1_cpr0_1PIs_Width} and \ref{fig:ApproxNeuraLCB_cp_epochs100_buffer-1_bs32_lr0_0001_beta1_0_std0_1_B10_G1_PIs_Width} illustrate the average width of prediction intervals for the two algorithms. For both methods, interval widths decrease initially but stabilize as the training size increases. NeuroSep-CP-LCB provides narrower intervals than NeuroSep-CP-LinLCB in the long run, particularly for higher $\alpha$ values. 

\paragraph{Linear Model Simplicity (NeuroSep-CP-LinLCB):}
By design, the linear model in NeuroSep-CP-LinLCB imposes fewer parameters and a simpler structure than a neural network. This simplicity allows for faster convergence in the early stages of training, yielding tighter prediction intervals even with limited data.
Despite its early advantages, the linear model's simplicity makes it less capable of capturing complex patterns in the data, particularly as the dataset grows. This limitation prevents it from refining its prediction intervals as effectively as the more flexible neural network in NeuroSep-CP-LCB.
\paragraph{Impact of Dataset Homogeneity:}
The reward function variability is inherently lower in homogeneous datasets, such as the septic-only dataset in this context. This enables linear models to perform well in the early stages by fitting straightforward patterns.
Neural networks, while more powerful in capturing nuanced patterns, initially struggle due to their higher capacity. However, as more data is seen, the neural model's ability to represent complex patterns allows it to surpass the linear model in prediction refinement.

In conclusion, the observed difference in average width between NeuroSep-CP-LinLCB and NeuroSep-CP-LCB can be explained by the fundamental trade-offs between model complexity and training data requirements. While NeuroSep-CP-LCB starts with more variability due to the inherent nature of neural networks and their sensitivity to data scarcity, it eventually leverages its capacity for greater flexibility to provide narrower and more accurate intervals in the long run. Conversely, NeuroSep-CP-LinLCB benefits from faster initial convergence but plateaus as the dataset grows, constrained by its linear assumptions and limited ability to model complex patterns.
 
\ref{fig:ApproxNeuralLinLCBJointModel_cp_epochs100_buffer-1_bs32_lr0_0001_beta1_0_std0_1_B10_G1_cpr0_1PIs_Width} depicts the average width of the prediction intervals:
\begin{itemize}
    \item NeuroSep-CP-LinLCB produces narrower intervals compared to NeuroSep-CP-LCB, particularly for higher $\alpha$ values. For instance, at $\alpha=0.25$, the interval width stabilizes around $0.15$ when train size reaches $8,000$ and beyond, whereas NeuroSep-CP-LCB intervals are wider under similar settings.
    \item This narrower interval width indicates that NeuroSep-CP-LinLCB leverages the linear model's parametric assumptions to generate tighter bounds, improving the precision of its predictions without sacrificing coverage.
\end{itemize}

\subsubsection{Fluctuations in Early Training}
Compared to NeuroSep-CP-LCB, NeuroSep-CP-LinLCB exhibits fewer fluctuations in the early training stages. This behavior can be attributed to:
\begin{itemize}
    \item \textbf{Linear Model Initialization:} The linear component provides a strong prior for decision-making without sufficient data, reducing reliance on exploratory actions during the initial rounds.
    \item \textbf{Faster Confidence Stabilization:} By combining neural predictions with linear estimations, the algorithm balances exploration and exploitation more effectively, leading to earlier regret stabilization.
\end{itemize}

Overall, NeuroSep-CP-LinLCB demonstrates superior performance, particularly in early training stages and under settings requiring stable confidence bounds, validating its potential as a robust extension to NeuroSep-CP-LCB.

\section{Conclusion}
\label{sec:conc}
 \subsection{Discussion and Limitations}
The NeuroSep-CP-LCB and NeuroSep-CP-LinLCB algorithms present significant innovations for early sepsis prediction in ICU patients. By integrating conformal prediction with contextual bandit frameworks, these algorithms enable robust uncertainty quantification, enhancing the reliability of predictions in high-stakes clinical settings. NeuroSep-CP-LCB excels in capturing complex nonlinear patterns, particularly as training data increases, demonstrating its ability to minimize regret while providing reliable prediction intervals. On the other hand, NeuroSep-CP-LinLCB is designed for simplicity and faster convergence, making it well-suited for scenarios with limited computational resources or datasets with lower variability, such as septic-only data. Both approaches demonstrate the ability to balance exploration and exploitation effectively, but NeuroSep-CP-LCB's flexibility comes at the cost of higher computational complexity.

The hyperparameter sensitivity of both algorithms is evident, as their performance strongly depends on tuning parameters such as the learning rate (\(lr\)), confidence scaling parameter (\(\beta\)), and conformal significance level (\(\alpha\)). These parameters are critical in managing the trade-offs between exploration, prediction interval width, and coverage, highlighting the importance of context-specific tuning. Moreover, including prediction intervals offers clinicians actionable insights by providing an understanding of the reliability of model outputs. This is particularly valuable in early sepsis prediction, where delayed or inaccurate predictions can have severe consequences.
 
Despite their strengths, these algorithms have some limitations. Both frameworks rely on offline datasets collected via a behavior policy, which might not align with real-world data collection practices. Variability in data quality and missingness can further influence model performance. Although HyperImpute was used for data imputation, it is subject to biases, especially when missing data mechanisms are non-random. 

NeuroSep-CP-LCB, with its neural network backbone, demands substantial computational resources, potentially limiting its scalability in real-time or resource-constrained environments. The ensemble-based conformal prediction also adds computational overhead, particularly as the number of bootstrap models increases. In contrast, while NeuroSep-CP-LinLCB is faster and more interpretable, it may struggle in datasets with complex, nonlinear relationships, as its linear assumptions restrict its ability to model intricate patterns. Additionally, both methods require careful hyperparameter tuning, which can be challenging in real-world applications with limited labeled data or dynamic patient populations.

\subsection{Conclusion and Future Work}
The NeuroSep-CP-LCB and NeuroSep-CP-LinLCB algorithms advance the field of early sepsis prediction by combining contextual bandits, neural networks, and conformal prediction. They provide robust, interpretable, and reliable predictions, addressing the critical need for timely and accurate decision-making in clinical practice. While NeuroSep-CP-LCB is adept at capturing complex patterns, NeuroSep-CP-LinLCB offers a practical alternative for faster and more interpretable predictions in time-sensitive or resource-constrained scenarios. These findings underscore the adaptability of the NeuroSep framework and its potential for diverse clinical applications.
 
Future research can build on the strengths of these algorithms while addressing their limitations. Extending the frameworks to handle dynamic patient contexts more effectively, such as by incorporating temporal models like recurrent neural networks or transformers, could improve their ability to process sequential ICU data. Developing real-time implementations of these algorithms would further enable continuous monitoring and prediction in ICU environments, making them more practical for clinical use. 

The robustness of these algorithms to missing data could also be improved. Investigating more sophisticated imputation methods or incorporating uncertainty-aware models that explicitly account for missing data would mitigate biases introduced by imputation. Moreover, validating the algorithms on diverse multicenter datasets is essential to ensure generalizability across different patient populations and clinical settings. 

Enhancing explainability in NeuroSep-CP-LCB through attention mechanisms or feature attribution methods could make neural predictions more interpretable for clinicians. Hybrid frameworks that dynamically switch between NeuroSep-CP-LCB and NeuroSep-CP-LinLCB based on dataset complexity or computational constraints could offer the best of both approaches. By addressing these areas, future research could refine these algorithms further, paving the way for their deployment in clinical practice and ultimately improving patient outcomes in critical care settings.

\bibliographystyle{unsrt}  
\bibliography{references}  

\begin{thebibliography}{10}

\bibitem{yin2020asymptotically}
Ming Yin and Yu-Xiang Wang.
\newblock Asymptotically efficient off-policy evaluation for tabular reinforcement learning.
\newblock In {\em International Conference on Artificial Intelligence and Statistics}, pages 3948--3958. PMLR, 2020.

\bibitem{rashidinejad2021bridging}
Paria Rashidinejad, Banghua Zhu, Cong Ma, Jiantao Jiao, and Stuart Russell.
\newblock Bridging offline reinforcement learning and imitation learning: A tale of pessimism.
\newblock {\em Advances in Neural Information Processing Systems}, 34:11702--11716, 2021.

\bibitem{duan2020minimax}
Yaqi Duan, Zeyu Jia, and Mengdi Wang.
\newblock Minimax-optimal off-policy evaluation with linear function approximation.
\newblock In {\em International Conference on Machine Learning}, pages 2701--2709. PMLR, 2020.

\bibitem{jin2021pessimism}
Ying Jin, Zhuoran Yang, and Zhaoran Wang.
\newblock Is pessimism provably efficient for offline rl?
\newblock In {\em International Conference on Machine Learning}, pages 5084--5096. PMLR, 2021.

\bibitem{nguyen2021sample}
Thanh Nguyen-Tang, Sunil Gupta, Hung Tran-The, and Svetha Venkatesh.
\newblock Sample complexity of offline reinforcement learning with deep relu networks.
\newblock {\em arXiv preprint arXiv:2103.06671}, 2021.

\bibitem{david2021offline}
Brandfonbrener David, F~Whitney William, Ranganath Rajesh, and Bruna Joan.
\newblock Offline contextual bandits with overparametrised models.
\newblock In {\em international conference on machine learning}, 2021.

\bibitem{allen2019convergence}
Zeyuan Allen-Zhu, Yuanzhi Li, and Zhao Song.
\newblock A convergence theory for deep learning via over-parameterization.
\newblock In {\em International conference on machine learning}, pages 242--252. PMLR, 2019.

\bibitem{cao2019generalization}
Yuan Cao and Quanquan Gu.
\newblock Generalization bounds of stochastic gradient descent for wide and deep neural networks.
\newblock {\em Advances in neural information processing systems}, 32, 2019.

\bibitem{nguyen-tang2022offline}
Thanh Nguyen-Tang, Sunil Gupta, A.~Tuan Nguyen, and Svetha Venkatesh.
\newblock Offline neural contextual bandits: Pessimism, optimization and generalization.
\newblock In {\em International Conference on Learning Representations}, 2022.

\bibitem{conformaltutorial}
Glenn Shafer and Vladimir Vovk.
\newblock A tutorial on conformal prediction.
\newblock {\em Journal of Machine Learning Research}, 9(Mar):371--421, 2008.

\bibitem{angelopoulos2021gentle}
Anastasios~N Angelopoulos and Stephen Bates.
\newblock A gentle introduction to conformal prediction and distribution-free uncertainty quantification.
\newblock {\em arXiv preprint arXiv:2107.07511}, 2021.

\bibitem{tibshirani2019conformal}
Ryan~J Tibshirani, Rina Foygel~Barber, Emmanuel Candes, and Aaditya Ramdas.
\newblock Conformal prediction under covariate shift.
\newblock {\em Advances in neural information processing systems}, 32, 2019.

\bibitem{lei2021conformal}
Lihua Lei and Emmanuel~J Cand{\`e}s.
\newblock Conformal inference of counterfactuals and individual treatment effects.
\newblock {\em Journal of the Royal Statistical Society Series B: Statistical Methodology}, 83(5):911--938, 2021.

\bibitem{abbasi2011improved}
Yasin Abbasi-Yadkori, D{\'a}vid P{\'a}l, and Csaba Szepesv{\'a}ri.
\newblock Improved algorithms for linear stochastic bandits.
\newblock {\em Advances in neural information processing systems}, 24, 2011.

\bibitem{zhou2020neural}
Dongruo Zhou, Lihong Li, and Quanquan Gu.
\newblock Neural contextual bandits with ucb-based exploration.
\newblock In {\em International Conference on Machine Learning}, pages 11492--11502. PMLR, 2020.

\bibitem{xiao2021optimality}
Chenjun Xiao, Yifan Wu, Jincheng Mei, Bo~Dai, Tor Lattimore, Lihong Li, Csaba Szepesvari, and Dale Schuurmans.
\newblock On the optimality of batch policy optimization algorithms.
\newblock In {\em International Conference on Machine Learning}, pages 11362--11371. PMLR, 2021.

\bibitem{jacot2018neuralNTK}
Arthur Jacot, Franck Gabriel, and Cl{\'e}ment Hongler.
\newblock Neural tangent kernel: Convergence and generalization in neural networks.
\newblock {\em Advances in neural information processing systems}, 31, 2018.

\bibitem{belkin2021fit}
Mikhail Belkin.
\newblock Fit without fear: remarkable mathematical phenomena of deep learning through the prism of interpolation.
\newblock {\em Acta Numerica}, 30:203--248, 2021.

\bibitem{hanin2019finite}
Boris Hanin and Mihai Nica.
\newblock Finite depth and width corrections to the neural tangent kernel.
\newblock {\em arXiv preprint arXiv:1909.05989}, 2019.

\bibitem{arora2019exact}
Sanjeev Arora, Simon~S Du, Wei Hu, Zhiyuan Li, Russ~R Salakhutdinov, and Ruosong Wang.
\newblock On exact computation with an infinitely wide neural net.
\newblock {\em Advances in neural information processing systems}, 32, 2019.

\bibitem{jacot2018neural}
Arthur Jacot, Franck Gabriel, and Cl{\'e}ment Hongler.
\newblock Neural tangent kernel: Convergence and generalization in neural networks.
\newblock {\em Advances in neural information processing systems}, 31, 2018.

\bibitem{du2018gradient}
Simon~S Du, Xiyu Zhai, Barnabas Poczos, and Aarti Singh.
\newblock Gradient descent provably optimizes over-parameterized neural networks.
\newblock {\em arXiv preprint arXiv:1810.02054}, 2018.

\bibitem{du2019gradient}
Simon Du, Jason Lee, Haochuan Li, Liwei Wang, and Xiyu Zhai.
\newblock Gradient descent finds global minima of deep neural networks.
\newblock In {\em International conference on machine learning}, pages 1675--1685. PMLR, 2019.

\bibitem{jarrett2022hyperimpute}
Daniel Jarrett, Bogdan~C Cebere, Tennison Liu, Alicia Curth, and Mihaela van~der Schaar.
\newblock Hyperimpute: Generalized iterative imputation with automatic model selection.
\newblock In {\em International Conference on Machine Learning}, pages 9916--9937. PMLR, 2022.

\bibitem{riquelme2018deep}
Carlos Riquelme, George Tucker, and Jasper Snoek.
\newblock Deep bayesian bandits showdown: An empirical comparison of bayesian deep networks for thompson sampling.
\newblock {\em arXiv preprint arXiv:1802.09127}, 2018.

\bibitem{PACE}
PACE.
\newblock {\em {P}artnership for an {A}dvanced {C}omputing {E}nvironment ({PACE})}, 2017.

\end{thebibliography}

\section{Appendix}
\label{sec:appendix_neuralcb}

\begin{figure}[!ht]
\centering
\includegraphics[width=0.8\textwidth]{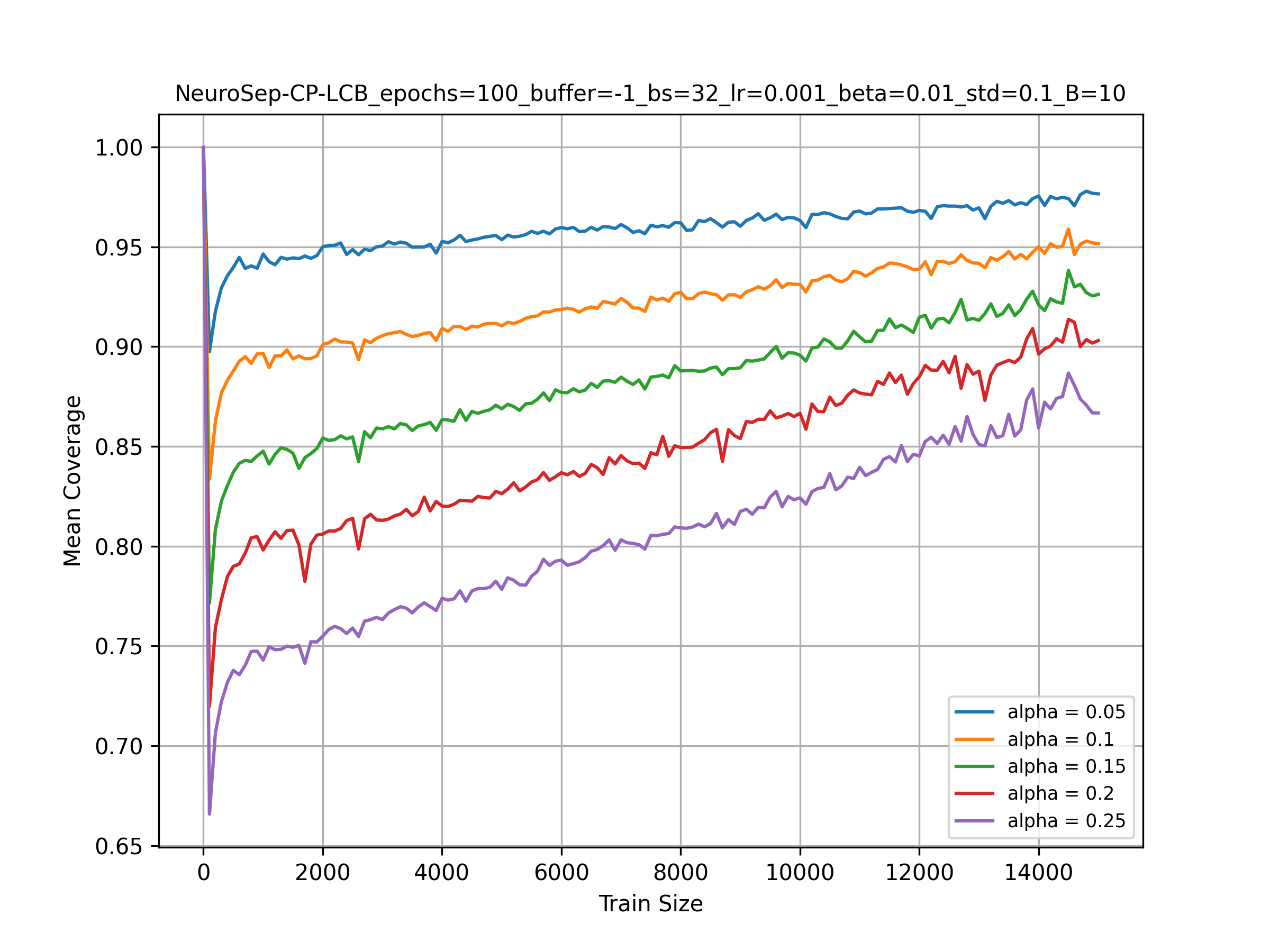}
\caption{Mean Coverage for NeuroSep-CP-LCB with lr=0.001, $\beta=0.01$.}
\label{fig:ApproxNeuraLCB_cp_epochs100_buffer-1_bs32_lr0_001_beta0_01_std0_1_B10_PIs_Coverage}
\end{figure}

\begin{figure}[!ht]
\centering
\includegraphics[width=0.8\textwidth]{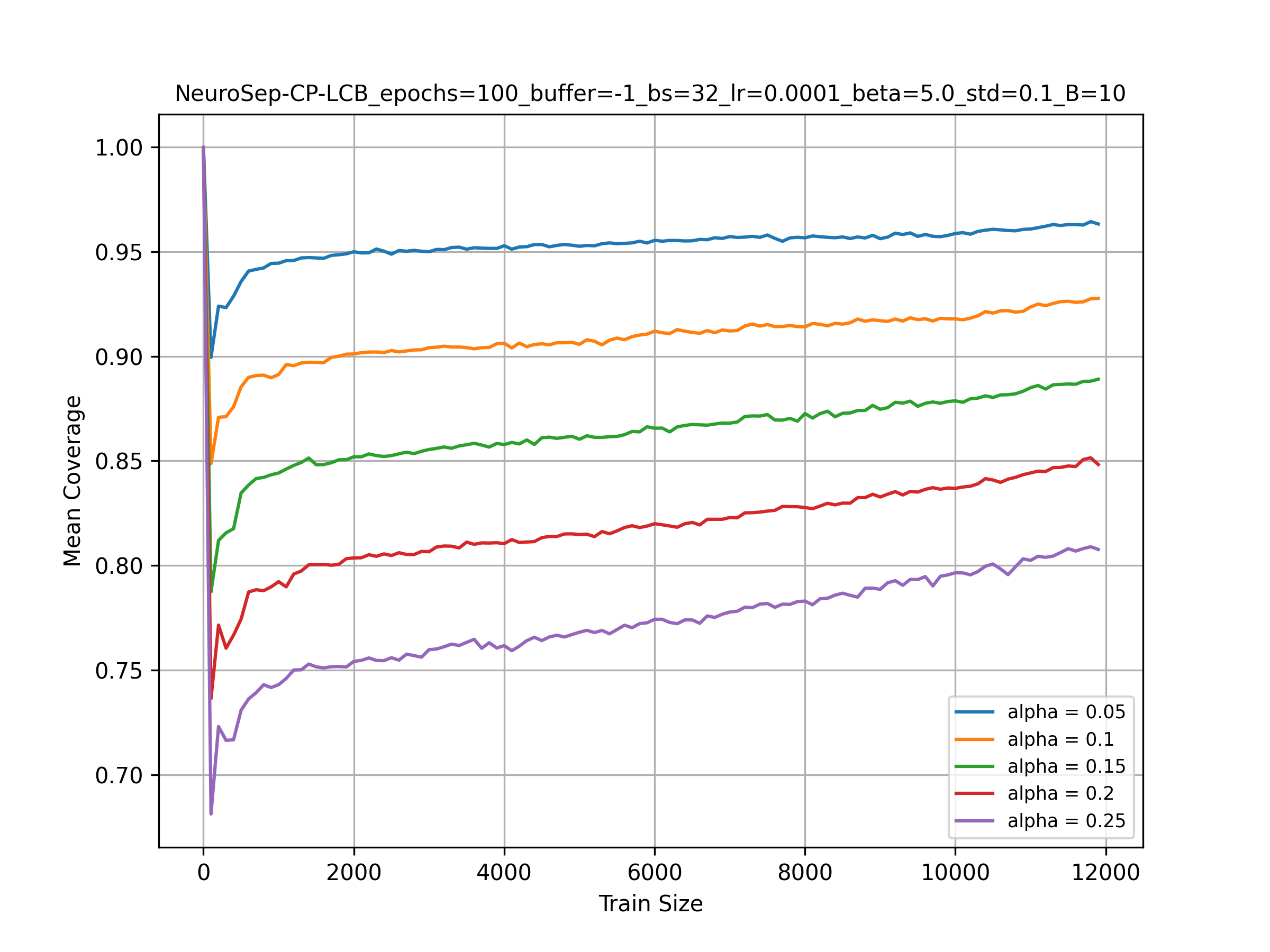}
\caption{Mean Coverage for NeuroSep-CP-LCB with lr=0.0001, $\beta=5.0$.}
\label{fig:ApproxNeuraLCB_cp_epochs100_buffer-1_bs32_lr0_0001_beta5_0_std0_1_B10_PIs_Coverage}
\end{figure}

\begin{figure}[!ht]
\centering
\includegraphics[width=0.8\textwidth]{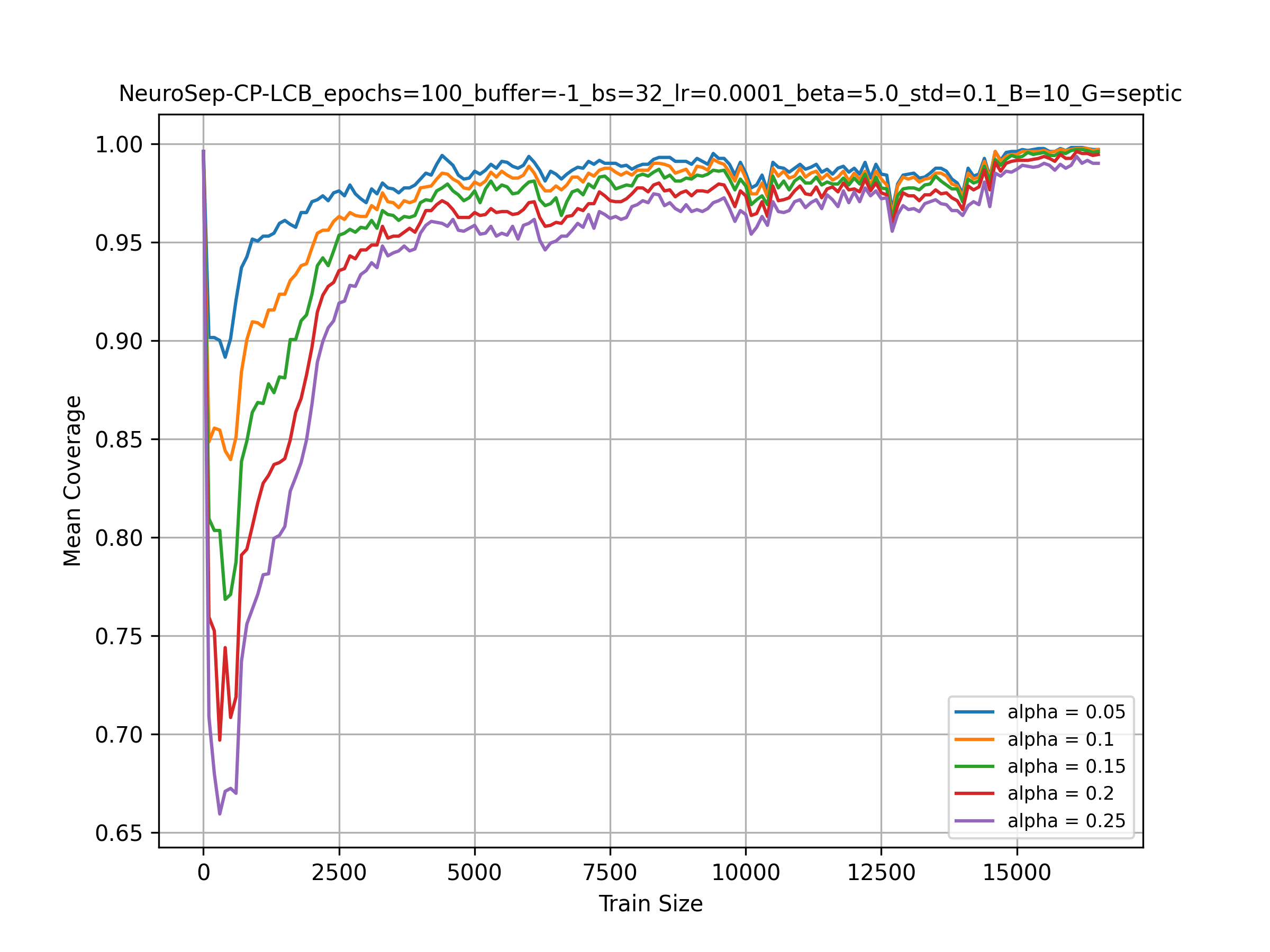}
\caption{Mean Coverage for NeuroSep-CP-LCB with lr=0.0001, $\beta=5.0$.}
\label{fig:ApproxNeuraLCB_cp_epochs100_buffer-1_bs32_lr0_0001_beta5_0_std0_1_B10_G1_PIs_Coverage}
\end{figure}


\begin{figure}[!ht]
\centering
\includegraphics[width=0.8\textwidth]{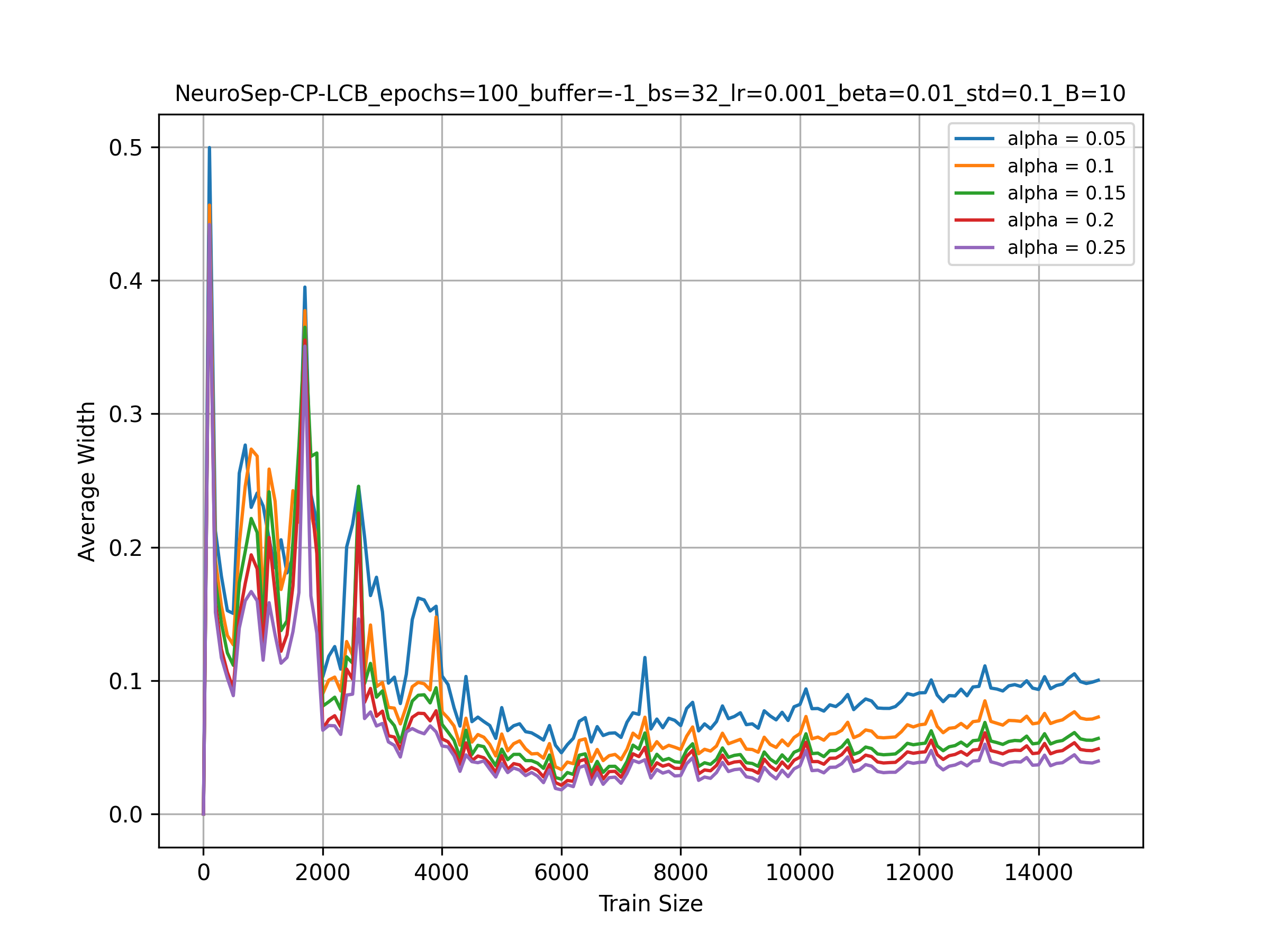}
\caption{Average Width for NeuroSep-CP-LCB with lr=0.001, $\beta=0.01$.}
\label{fig:ApproxNeuraLCB_cp_epochs100_buffer-1_bs32_lr0_001_beta0_01_std0_1_B10_PIs_Width}
\end{figure}

\begin{figure}[!ht]
\centering
\includegraphics[width=0.8\textwidth]{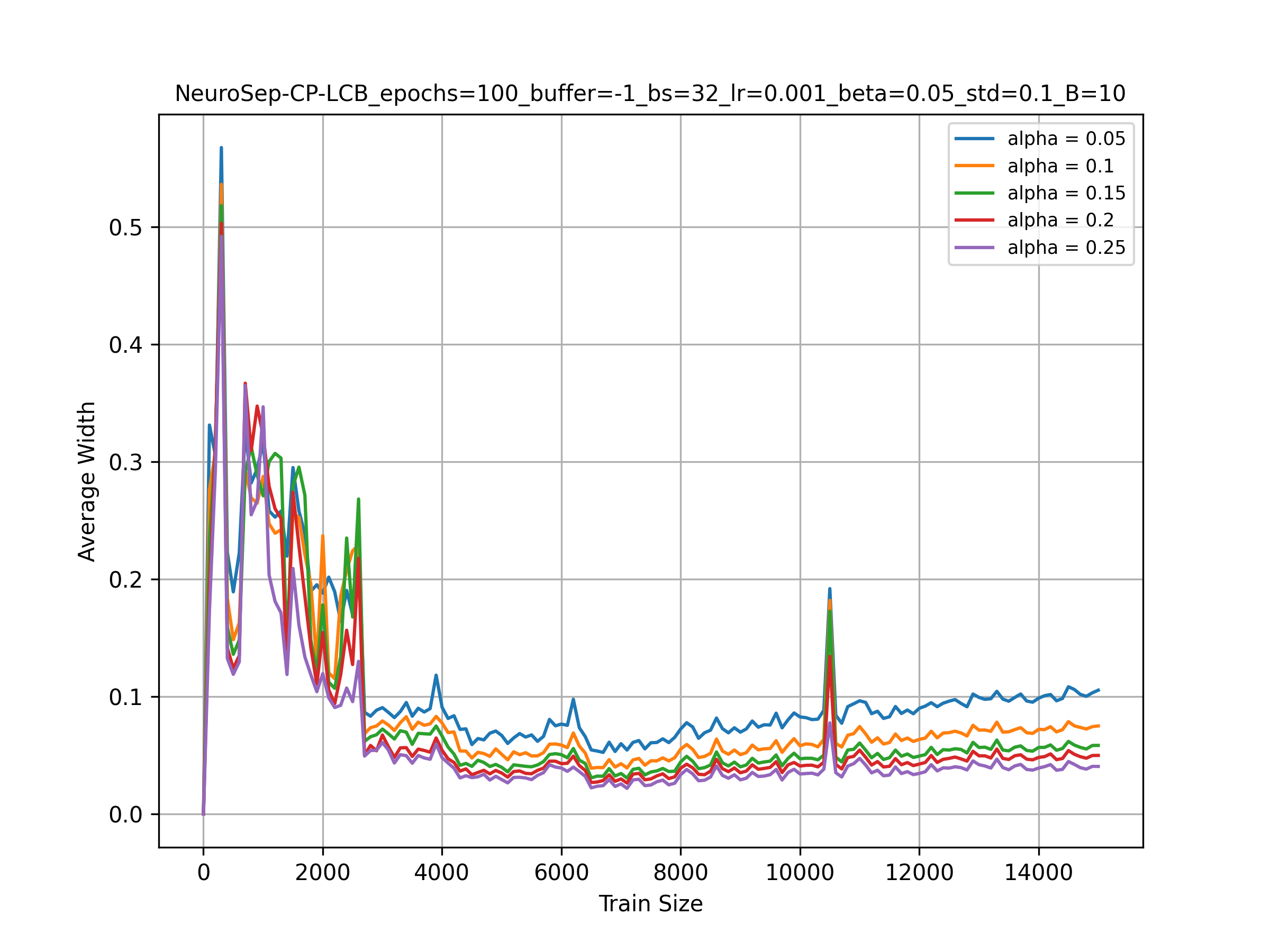}
\caption{Average Width for NeuroSep-CP-LCB with lr=0.001, $\beta=0.05$.}
\label{fig:ApproxNeuraLCB_cp_epochs100_buffer-1_bs32_lr0_001_beta0_05_std0_1_B10_PIs_Width}
\end{figure}

\begin{figure}[!ht]
\centering
\includegraphics[width=0.8\textwidth]{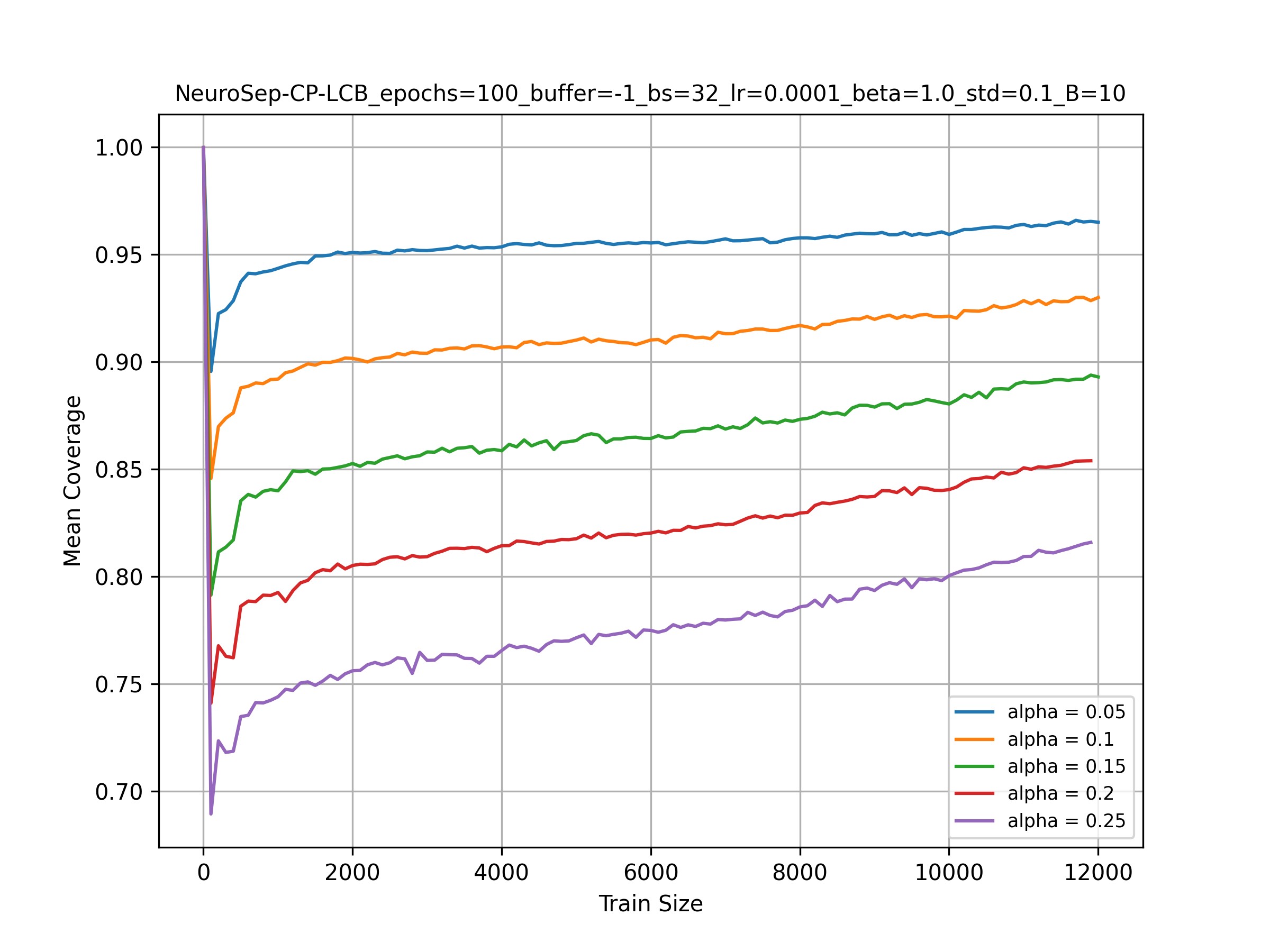}
\caption{Mean Coverage for NeuroSep-CP-LCB with lr=0.0001, $\beta=1.0$.}
\label{fig:ApproxNeuraLCB_cp_epochs100_buffer-1_bs32_lr0_0001_beta1_0_std0_1_B10_PIs_Coverage}
\end{figure}

\begin{figure}[!ht]
\centering
\includegraphics[width=0.8\textwidth]{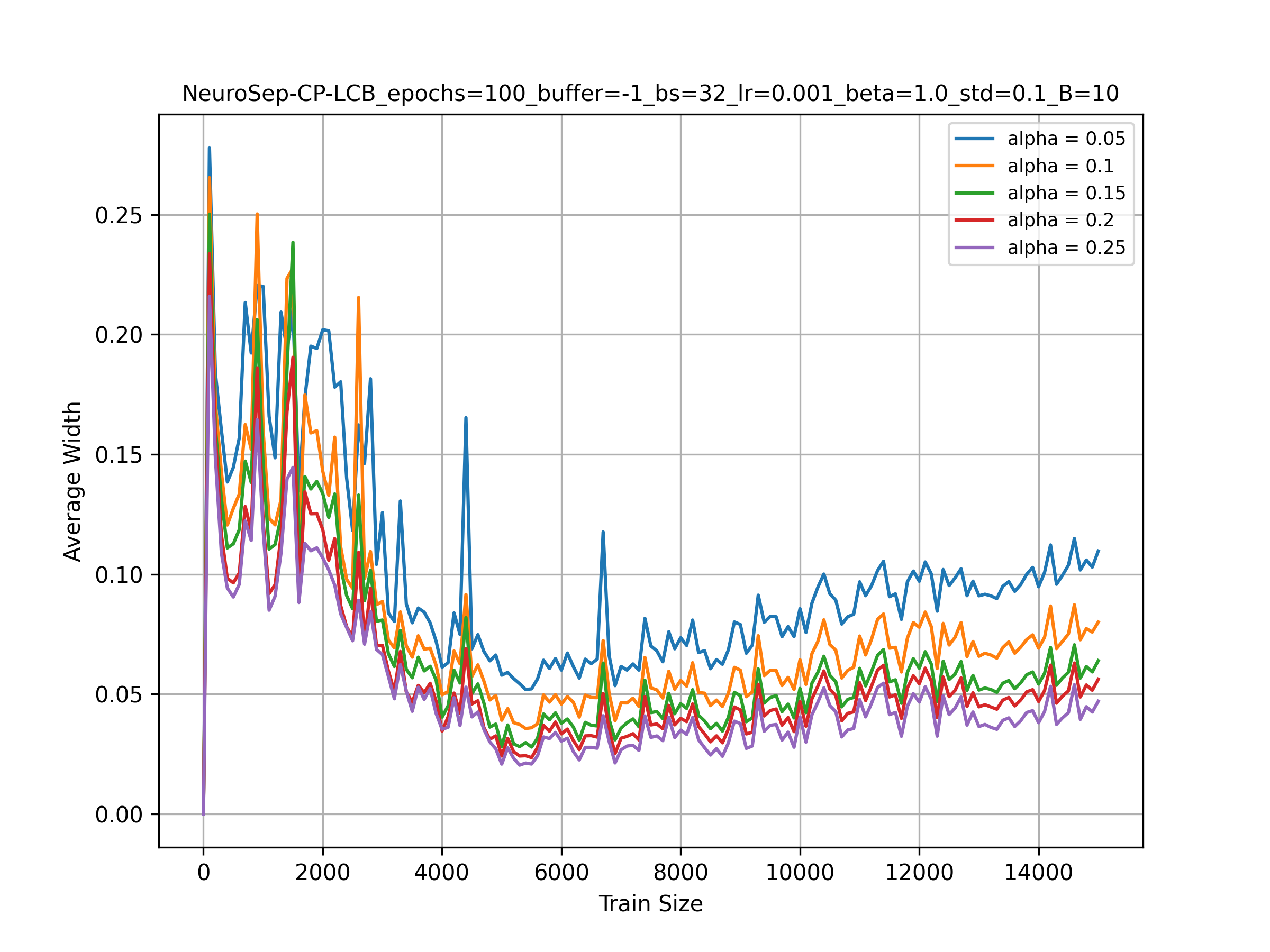}
\caption{Average Width for NeuroSep-CP-LCB with lr=0.001, $\beta=1.0$.}
\label{fig:ApproxNeuraLCB_cp_epochs100_buffer-1_bs32_lr0_001_beta1_0_std0_1_B10_PIs_Width}
\end{figure}

\begin{figure}[!ht]
\centering
\includegraphics[width=0.8\textwidth]{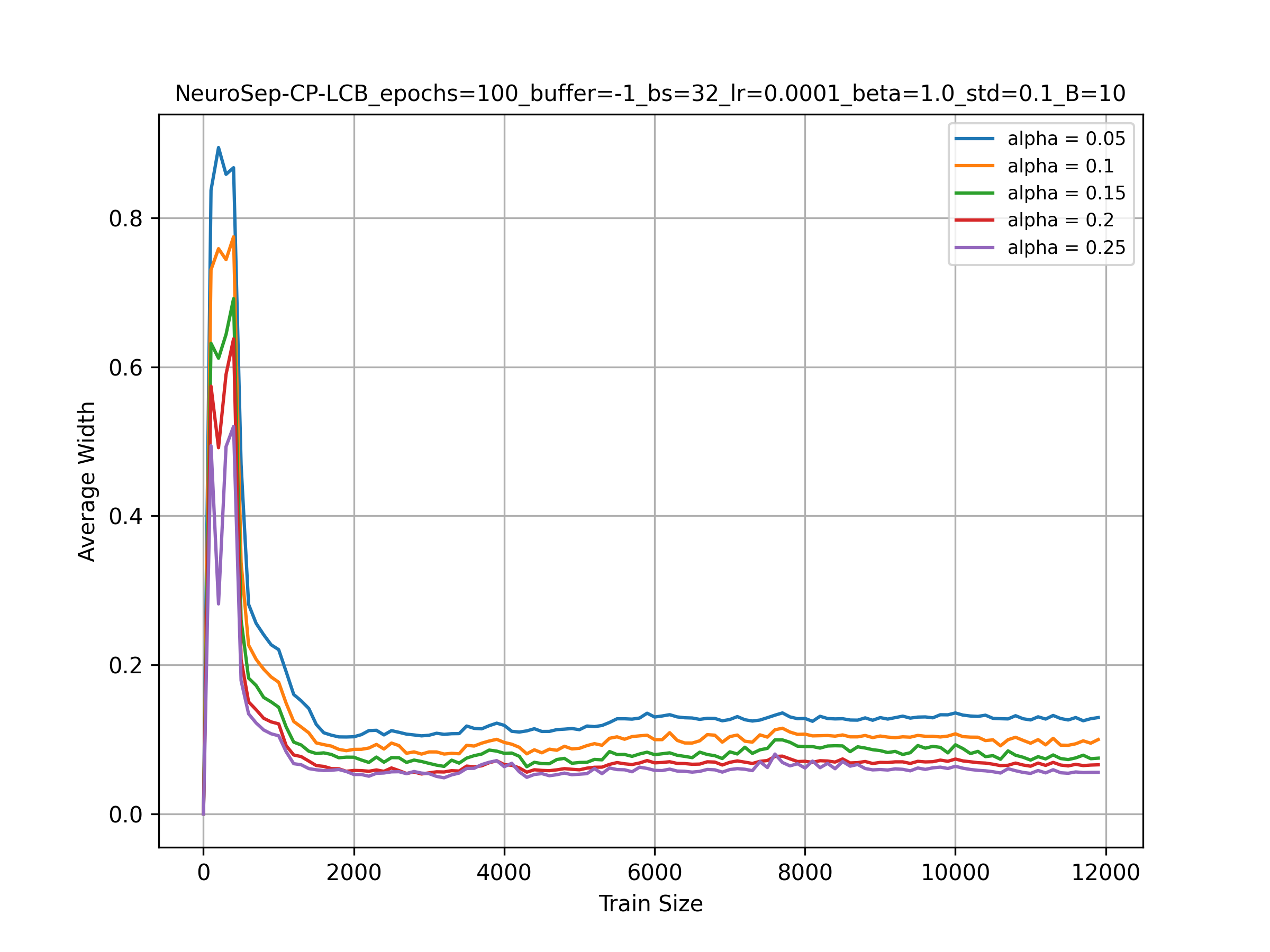}
\caption{Average Width for NeuroSep-CP-LCB with lr=0.0001, $\beta=1.0$.}
\label{fig:ApproxNeuraLCB_cp_epochs100_buffer-1_bs32_lr0_0001_beta1_0_std0_1_B10_PIs_Width}
\end{figure}

\begin{figure}[!ht]
\centering
\includegraphics[width=0.8\textwidth]{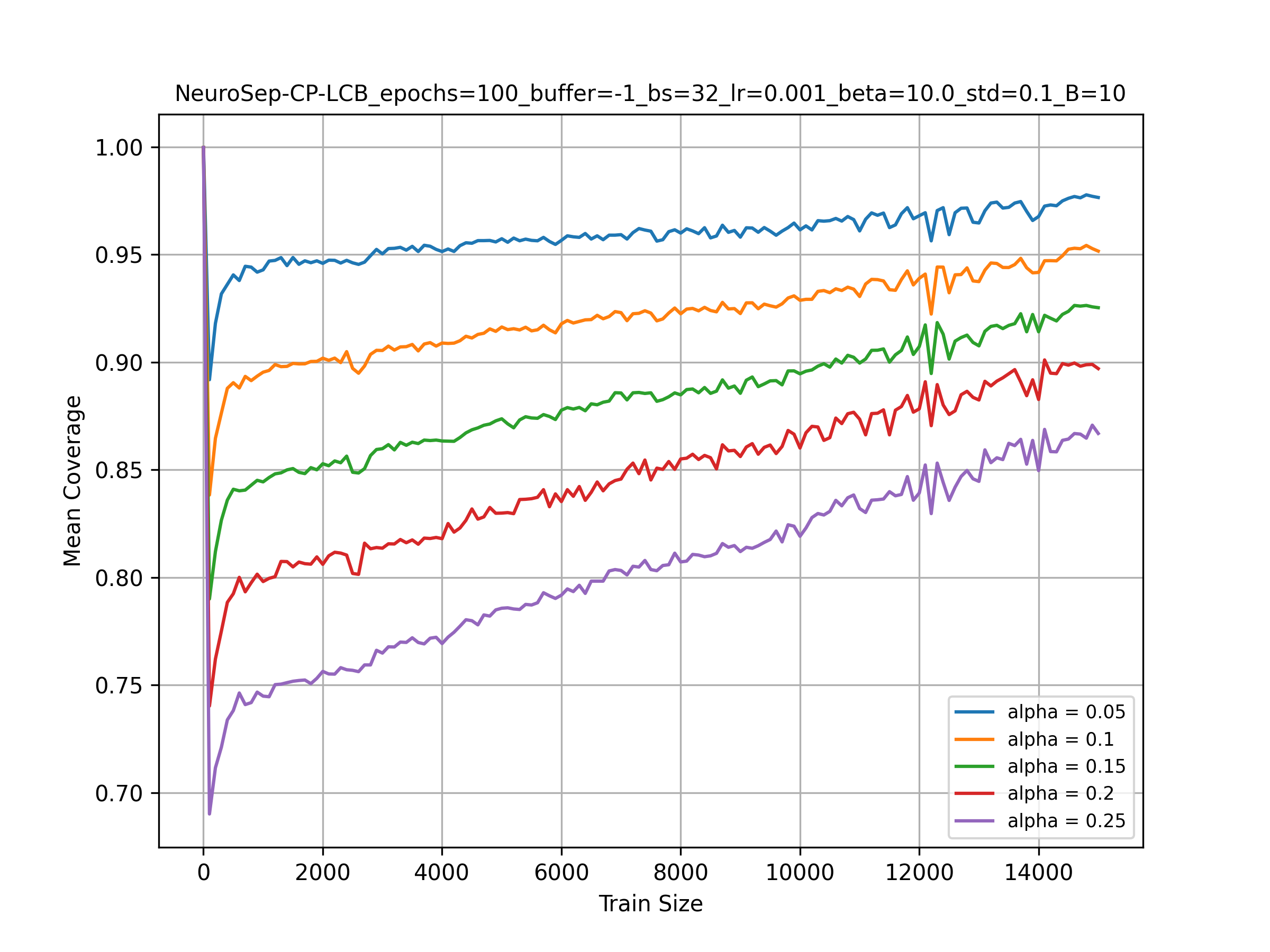}
\caption{Mean Coverage for NeuroSep-CP-LCB with lr=0.001, $\beta=10.0$.}
\label{fig:ApproxNeuraLCB_cp_epochs100_buffer-1_bs32_lr0_001_beta10_0_std0_1_B10_PIs_Coverage}
\end{figure}

\begin{figure}[!ht]
\centering
\includegraphics[width=0.8\textwidth]{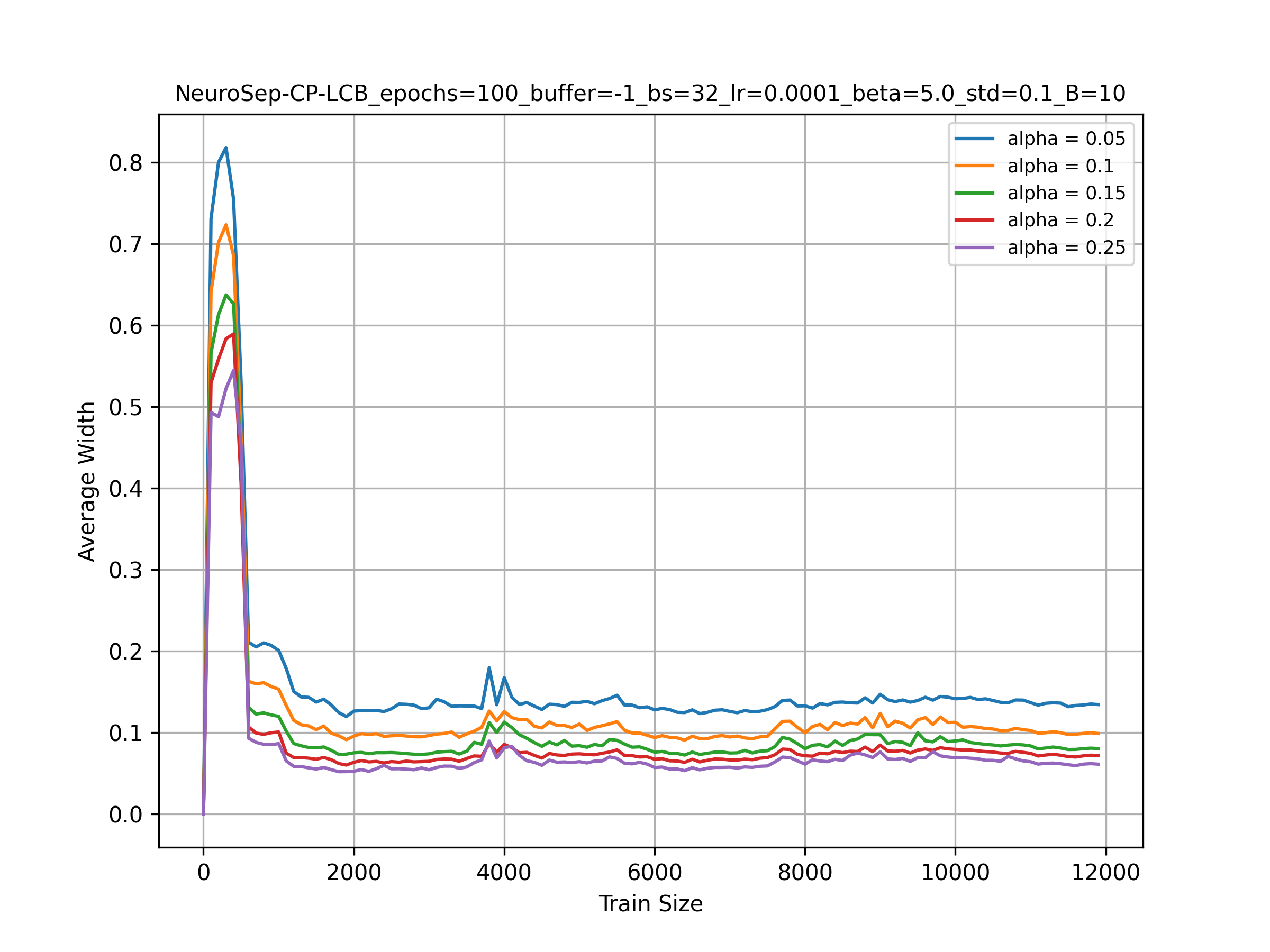}
\caption{Average Width for NeuroSep-CP-LCB with lr=0.0001, $\beta=5.0$.}
\label{fig:ApproxNeuraLCB_cp_epochs100_buffer-1_bs32_lr0_0001_beta5_0_std0_1_B10_PIs_Width}
\end{figure}

\begin{figure}[!ht]
\centering
\includegraphics[width=0.8\textwidth]{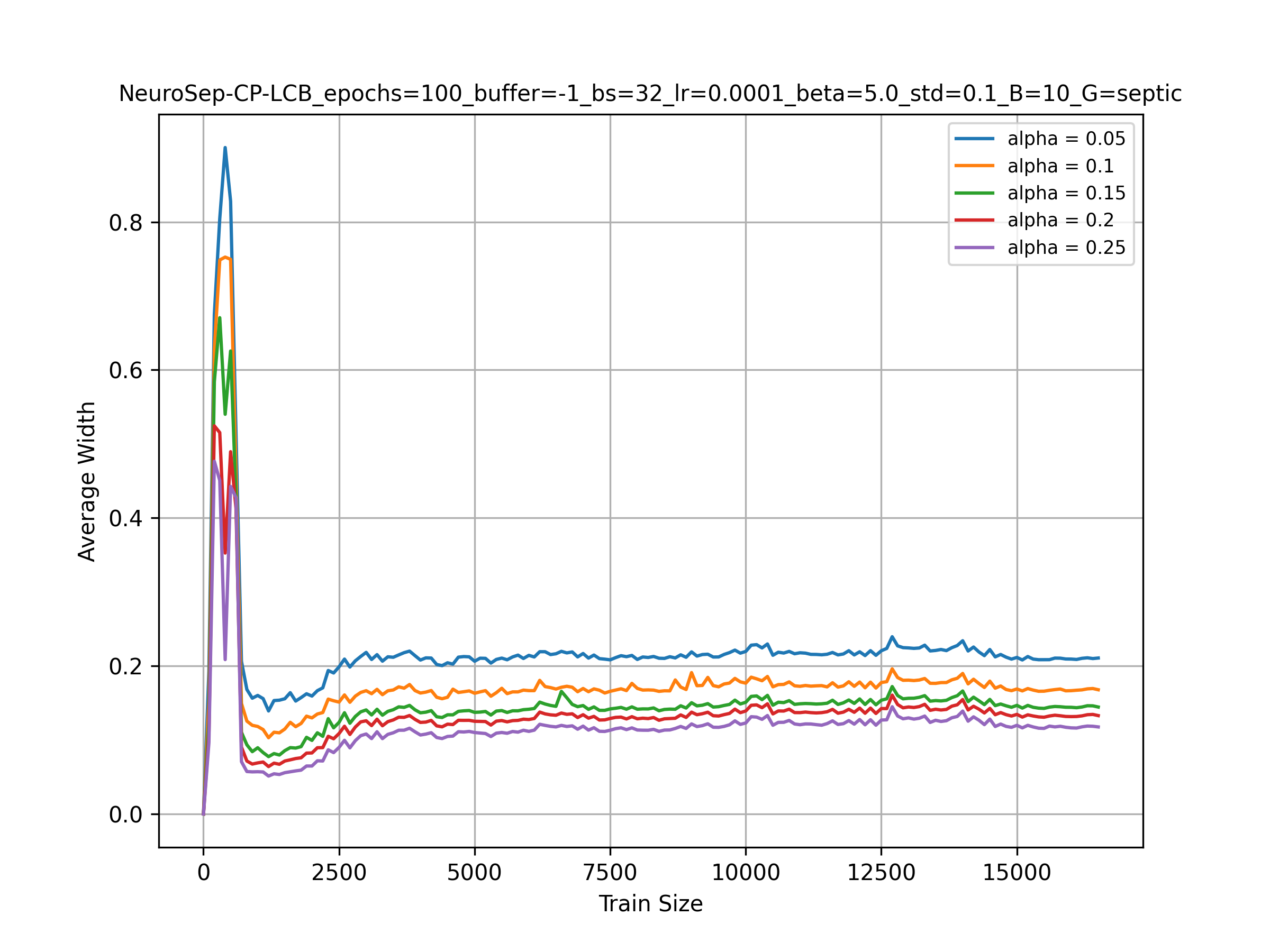}
\caption{Average Width for NeuroSep-CP-LCB with lr=0.0001, $\beta=5.0$.}
\label{fig:ApproxNeuraLCB_cp_epochs100_buffer-1_bs32_lr0_0001_beta5_0_std0_1_B10_G1_PIs_Width}
\end{figure}

\begin{figure}[!ht]
\centering
\includegraphics[width=0.8\textwidth]{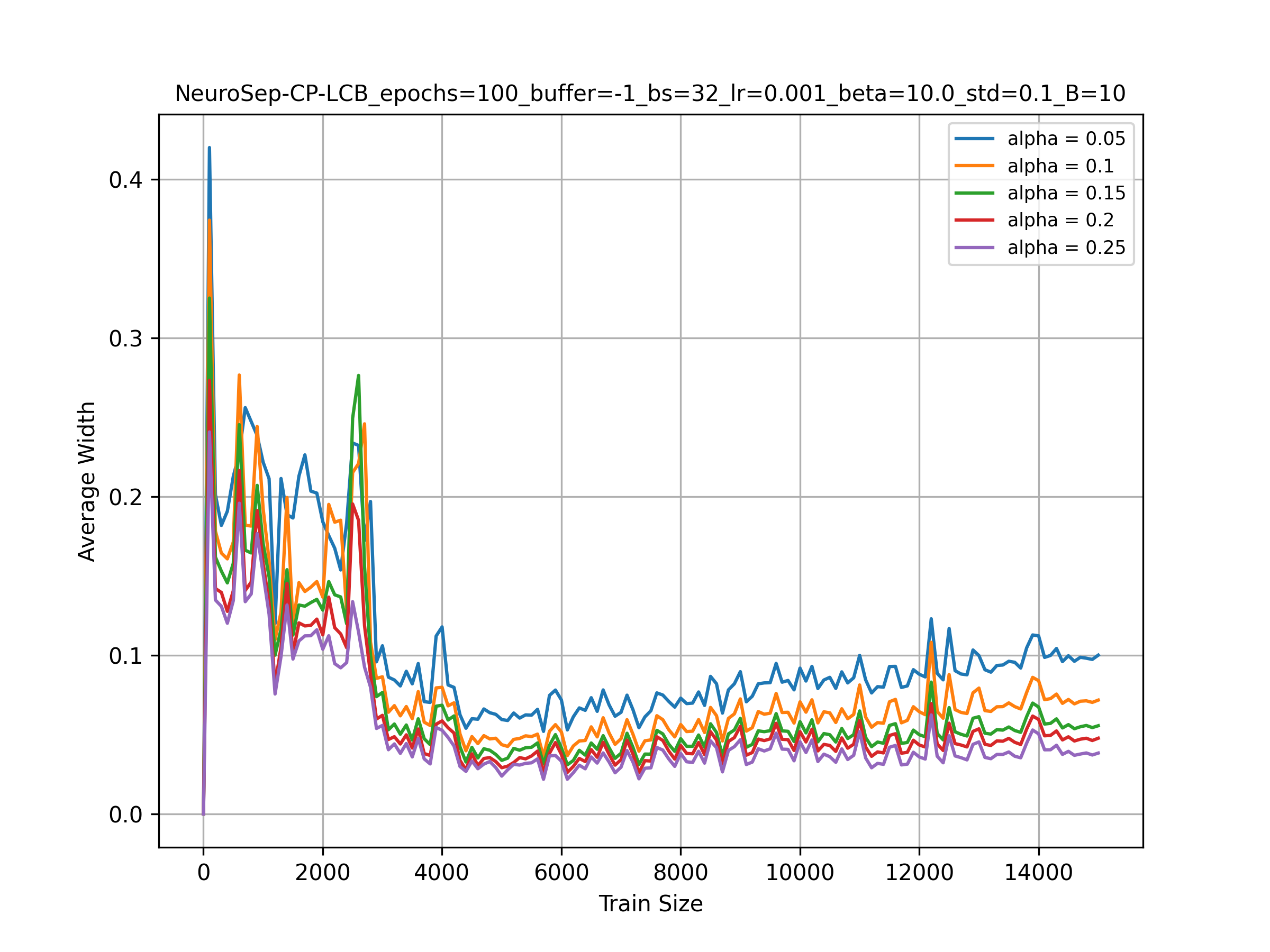}
\caption{Average Width for NeuroSep-CP-LCB with lr=0.001, $\beta=10.0$.}
\label{fig:ApproxNeuraLCB_cp_epochs100_buffer-1_bs32_lr0_001_beta10_0_std0_1_B10_PIs_Width}
\end{figure}


\begin{figure}[!ht]
\centering
\includegraphics[width=0.8\textwidth]{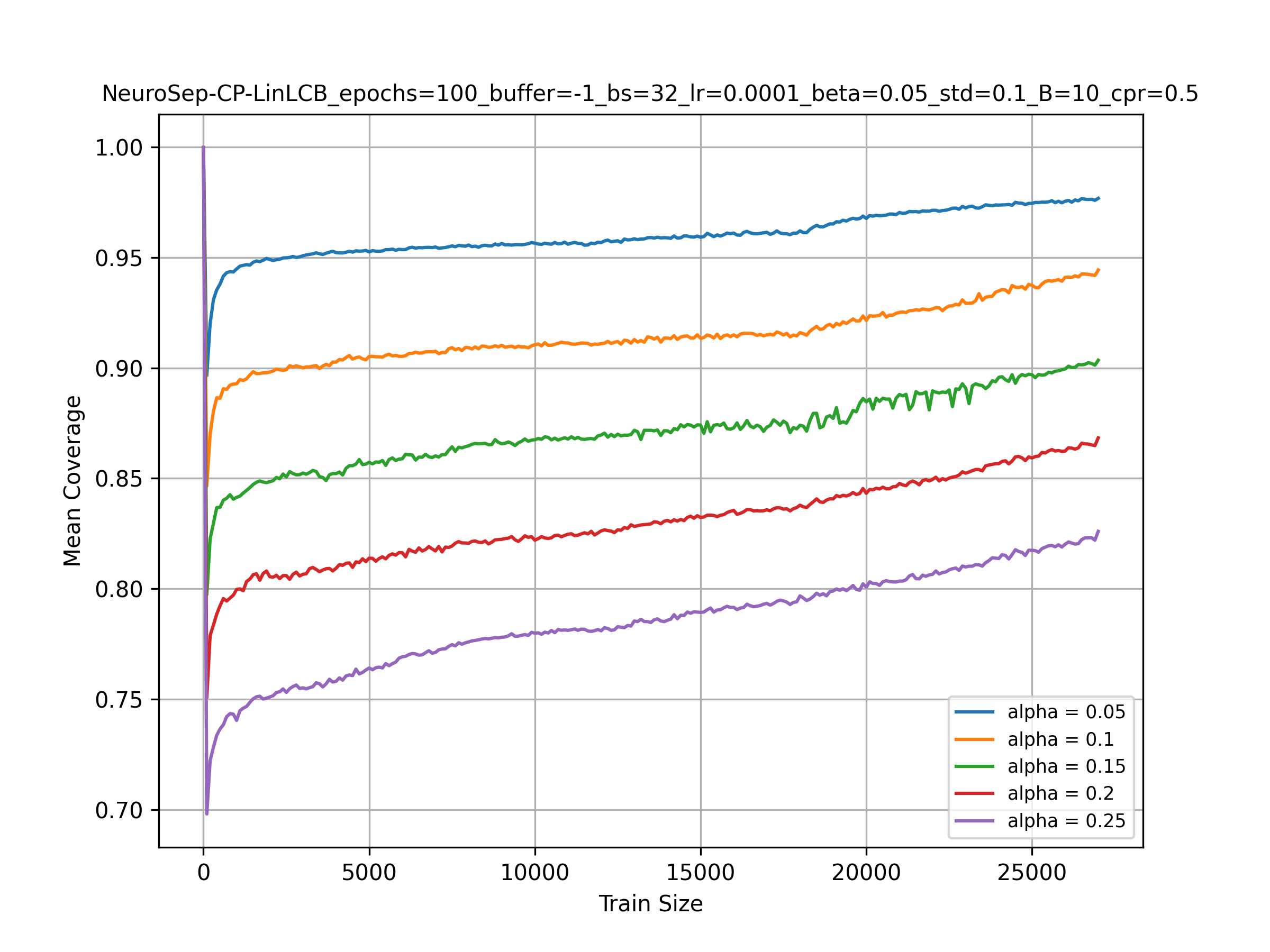}
\caption{Mean Coverage for NeuroSep-CP-LinLCB with lr=0.0001, $\beta=0.05$, batch size=32.}
\label{fig:ApproxNeuralLinLCBJointModel_cp_epochs100_buffer-1_bs32_lr0_0001_beta0_05_std0_1_B10_cpr0_5PIs_Coverage}
\end{figure}

\begin{figure}[!ht]
\centering
\includegraphics[width=0.8\textwidth]{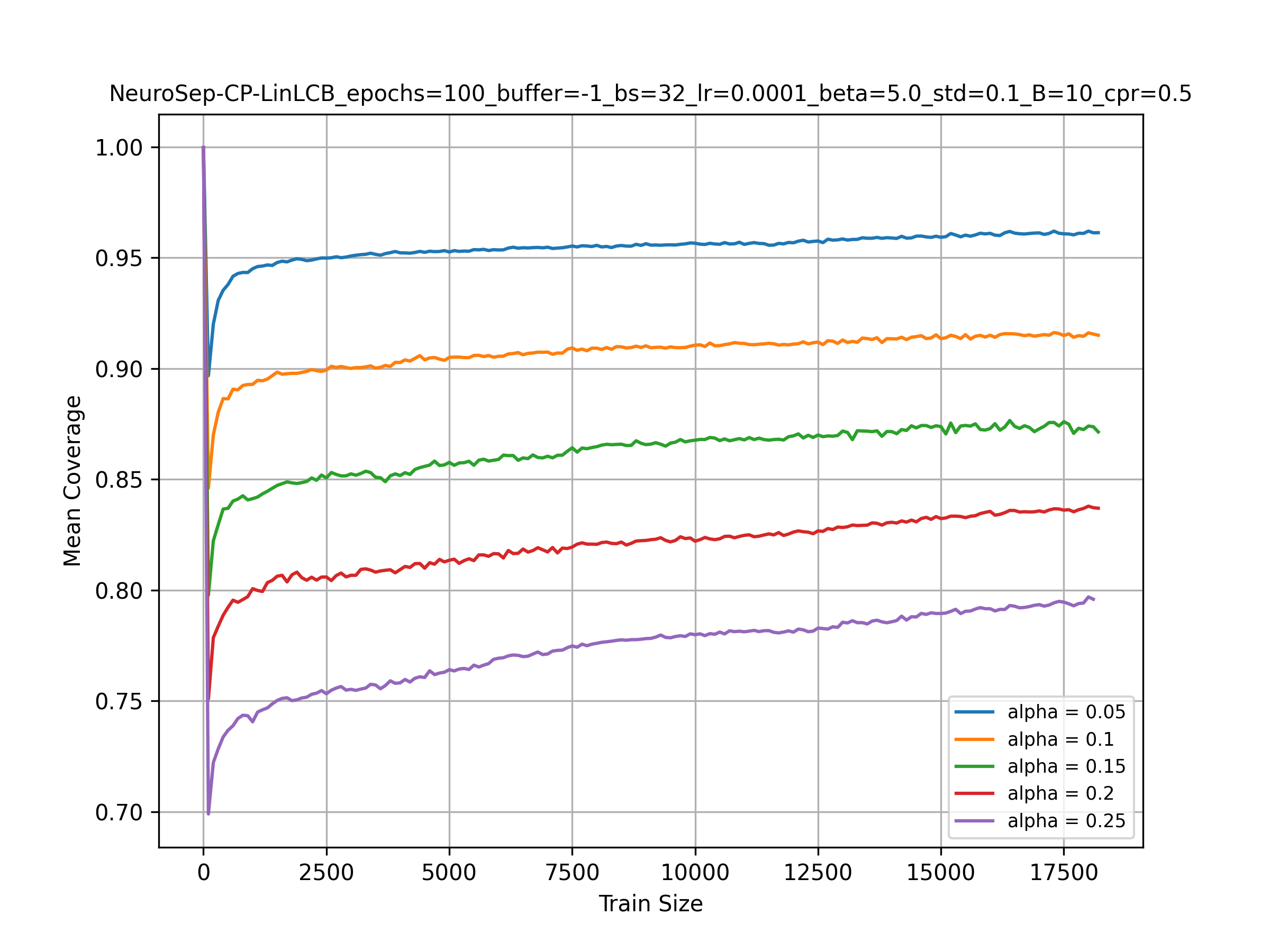}
\caption{Mean Coverage for NeuroSep-CP-LinLCB with lr=0.0001, $\beta=5.0$.}
\label{fig:ApproxNeuralLinLCBJointModel_cp_epochs100_buffer-1_bs32_lr0_0001_beta5_0_std0_1_B10_cpr0_5PIs_Coverage}
\end{figure}

\begin{figure}[!ht]
\centering
\includegraphics[width=0.8\textwidth]{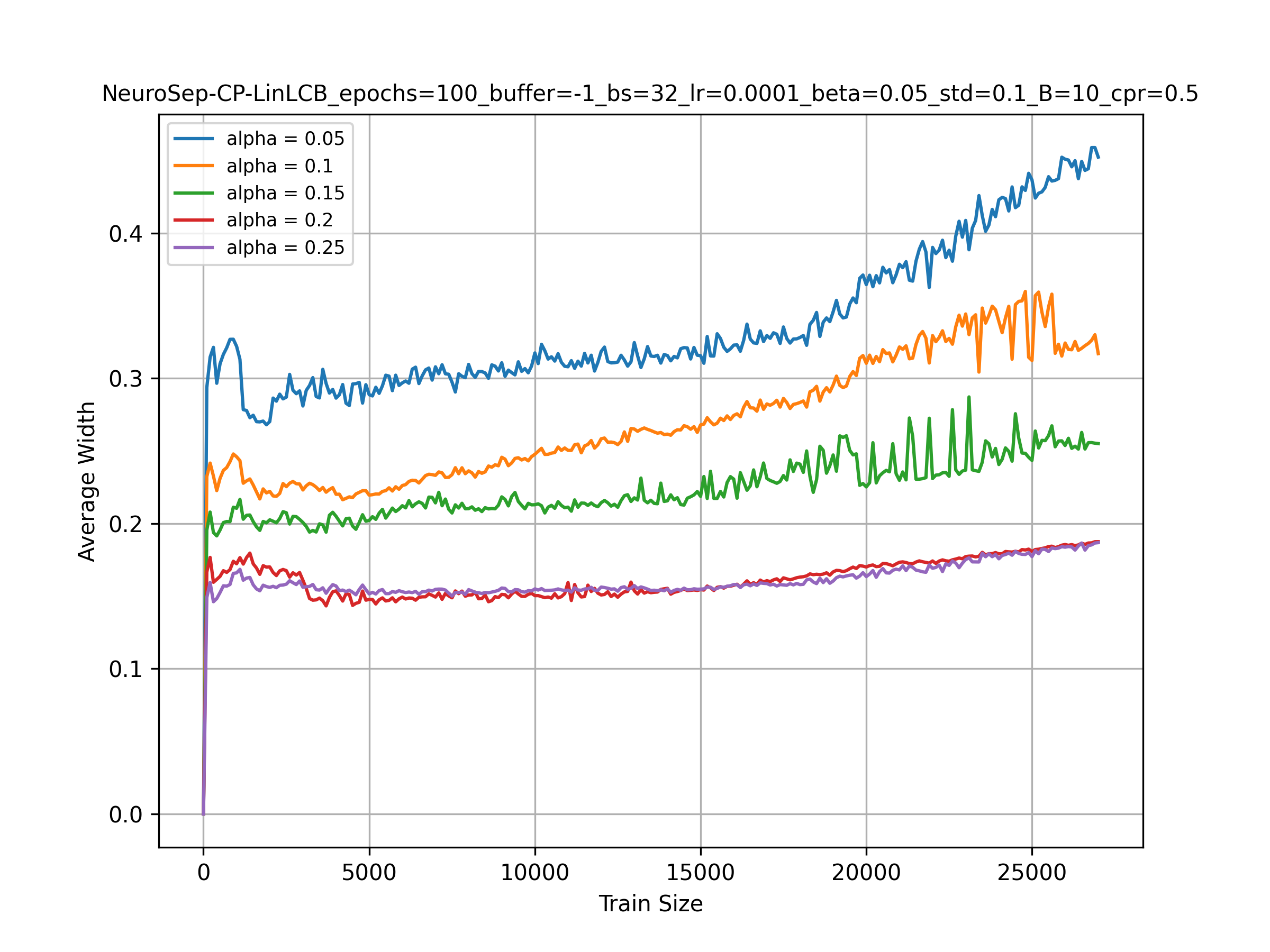}
\caption{Average Width for NeuroSep-CP-LinLCB with lr=0.0001, $\beta=0.05$.}
\label{fig:ApproxNeuralLinLCBJointModel_cp_epochs100_buffer-1_bs32_lr0_0001_beta0_05_std0_1_B10_cpr0_5PIs_Width}
\end{figure}

\begin{figure}[!ht]
\centering
\includegraphics[width=0.8\textwidth]{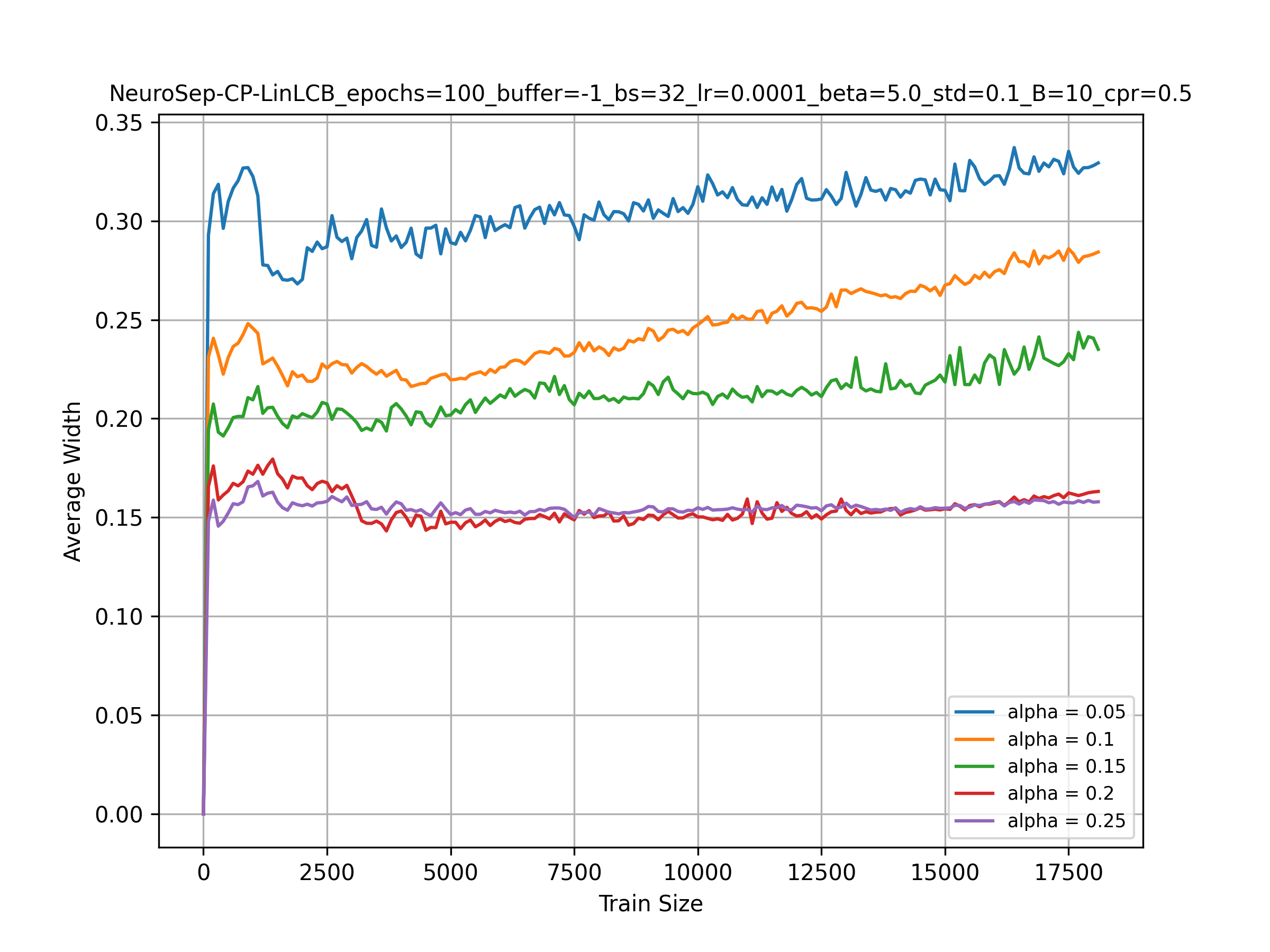}
\caption{Average Width for NeuroSep-CP-LinLCB with lr=0.0001, $\beta=5.0$.}
\label{fig:ApproxNeuralLinLCBJointModel_cp_epochs100_buffer-1_bs32_lr0_0001_beta5_0_std0_1_B10_cpr0_5PIs_Width}
\end{figure}

\end{document}